\documentclass[10pt,twocolumn,letterpaper]{article}
\usepackage[pagenumbers]{cvpr} 
\usepackage{amsthm}

\usepackage{xspace}
\usepackage{float}
\usepackage{multirow, booktabs, graphicx, colortbl, arydshln, tikz}
\usepackage[pagebackref,breaklinks,colorlinks,allcolors=cvprblue]{hyperref}
\definecolor{cvprblue}{rgb}{0.21,0.49,0.74}
\title{Towards Precise Scaling Laws for Video Diffusion Transformers}
\author{
Yuanyang Yin$^1$\footnotemark[1], Yaqi Zhao$^2$\footnotemark[1], Mingwu Zheng$^3$, Ke Lin$^3$\\
Jiarong Ou$^3$, Rui Chen$^3$, Victor Shea-Jay Huang$^2$, Jiahao Wang$^3$\\
Xin Tao$^3$, Pengfei Wan$^3$, Di Zhang$^3$, Baoqun Yin$^1$\footnotemark[2], Wentao Zhang$^2$\footnotemark[2], Kun Gai$^3$\\
{$^1$University of Science and Technology of China\hspace{0.5cm}}
{$^2$Peking University\hspace{0.5cm}}
{$^3$Kuaishou Technology\hspace{0.5cm}}\\
}

\begin{document}
\maketitle
\renewcommand{\thefootnote}{\fnsymbol{footnote}}
\footnotetext[1]{Equal contribution.}
\footnotetext[2]{Corresponding Author.}
\begin{abstract}

Achieving optimal performance of video diffusion transformers within given data and compute budget is crucial due to their high training costs. This necessitates precisely determining the optimal model size and training hyperparameters before large-scale training. 
While scaling laws are employed in language models to predict performance, their existence and accurate derivation in visual generation models remain underexplored. 
In this paper, we systematically analyze scaling laws for video diffusion transformers and confirm their presence. Moreover, we discover that, unlike language models, video diffusion models are more sensitive to learning rate and batch size—two hyperparameters often not precisely modeled. To address this, we propose a new scaling law that predicts optimal hyperparameters for any model size and compute budget. 
Under these optimal settings, we achieve comparable performance and reduce inference costs by $40.1\%$ compared to conventional scaling methods, within a compute budget of $1e10$ TFlops. 
Furthermore, we establish a more generalized and precise relationship among validation loss, any model size, and compute budget. This enables performance prediction for non-optimal model sizes, which may also be appealed under practical inference cost constraints, achieving a better trade-off.
\end{abstract}    
\section{Introduction}\label{sec:intro}

\begin{figure*}[htbp]
    \centering
    \includegraphics[width=1.0\textwidth]{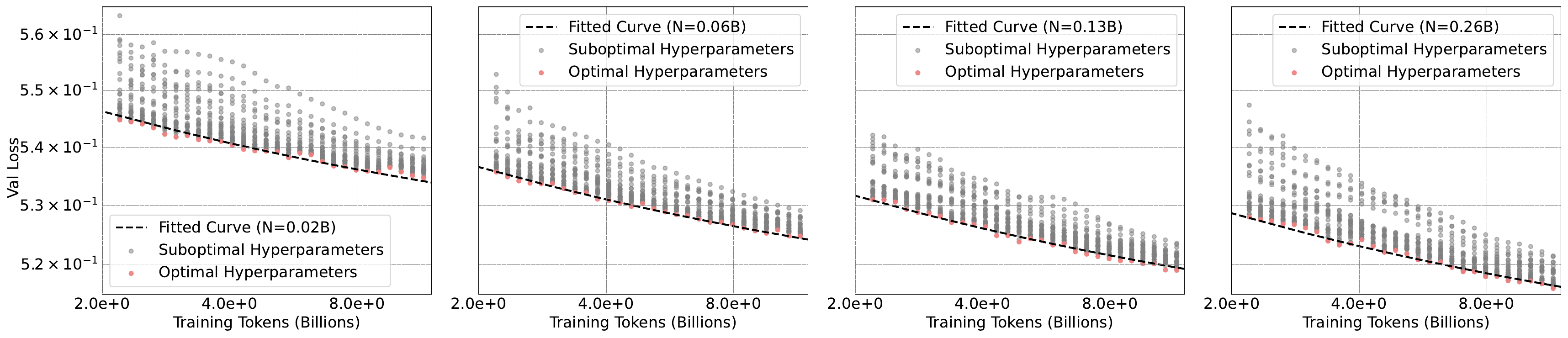}
    \caption{Validation loss across various model sizes with different amounts of training tokens and hyperparameters. Each panel represents a different model size (0.02B, 0.06B, 0.13B, and 0.26B). Gray points denote the validation loss achieved with fixed suboptimal hyperparameters, while red points highlight the lowest validation loss obtained with the optimal batch size and learning rate. This demonstrates that selecting optimal hyperparameters is essential for correctly fitting the loss curve, ensuring more accurate alignment with expected scaling trends as model size and data size increase.}
    \label{fig:SubOptimalHp}
\end{figure*}

Recent advancements in Diffusion Transformers (DiT)~\citep{ho2020denoising,song2020denoising,peebles2023scalable,ma2024sitexploringflowdiffusionbased} have significantly enhanced video generation capabilities over previous convolutional-based neural models~\citep{chen2024videocrafter2,girdhar2023emu,blattmann2023stable,ho2022imagen,singer2022make}, demonstrating superior quality and diversity in generated video content~\citep{opensora, lu2023vdt,ma2024latte,brooks2024video,gao2024lumina,zhang2024tora,klingai2024,sora2024}. The success of video DiT can be largely attributed to the scalability of transformer architecture, where simply scaling up the model size often yields improved performance. Currently, the largest video generation model like Movie Gen~\citep{polyak2024movie}, has reached a parameter count of 30 billion, which is significantly larger than the initial video DiT around 700M~\citep{ma2024latte, opensora}.

However, due to limited training budgets, infinitely scaling up transformer size is not feasible. Consequently, determining the optimal DiT size within a given budget has become a critical challenge.
Early research on transformer-based language models~\citep{henighan2020scaling,hoffmann2022training,kaplan2020scaling,bi2024deepseek,dubey2024llama} revealed that model performance (measured by validation loss $L$), the model size $N$, and the given compute budget $C$, exhibit a power-law relationship, known as \textit{scaling laws}. This discovery enables the use of smaller models for experiments to predict the performance of larger models, thereby enabling the selection of the optimal model size under real-world constrains.

In the field of computer vision, researchers have primarily made empirical attempts in the scaling properties~\citep{sauer2024fast, ma2024sitexploringflowdiffusionbased, peebles2023scalable, li2024scalability, mei2024bigger} of image DiT, seldom modeling precise scaling laws as done in language models. In the domain of video generation, such research is virtually nonexistent. Given the complexity and importance of video models—which must not only model visual structures but also simulate and understand complex real-world dynamics over time dimension—they are often trained at very high costs; for example, Movie Gen was trained using 6,144 H100 GPUs~\citep{polyak2024movie}. Therefore, developing accurate and practical scaling laws for large-scale video generation models is particularly important, as selecting the correct model size and training configurations can greatly reduce costs.

This study aims to investigate the scaling laws of video diffusion transformers by addressing the following key question: \textit{Given a specific video transformer architecture and training budget, how can we predict the optimal model size, along with its corresponding performance and optimal hyperparameters—specifically, batch size and learning rate?}

We began our investigation by applying a commonly used scaling law from large language models~\citep{hoffmann2022training} to video diffusion transformers. However, we found that due to the greater sensitivity of video diffusion models to hyperparameters such as batch size and learning rate, the predictions made using this scaling law were inaccurate. Relying on fixed, potentially suboptimal hyperparameters introduced substantial uncertainty and resulted in less precise scaling laws, as shown in~\Cref{fig:combine_Nopt,fig:SubOptimalHp}. When exceeding a small compute budget, this approach tend to select larger model sizes and lead to higher validation loss, as observed in ~\Cref{fig:Nopt_fixedNopt}. Therefore, identifying optimal hyperparameters in video diffusion transformers is essential. However, in the research on scaling laws for language models, the selection of optimal hyperparameters has often been overlooked or remains controversial~\citep{hoffmann2022training, kaplan2020scaling, bi2024deepseek, shen2024power}.

In this paper, we propose a new scaling law for the optimal hyperparameters to address these challenges. We show that selecting optimal hyperparameters depends primarily on model size and the number of training tokens. 
Based on this, we have obtained a more accurate prediction of the optimal model size under a given compute budget. By using the same computational resources as Movie Gen~\cite{polyak2024movie}, our approach allows us to reduce the model size by $39.9\%$ compared to predictions based on fixed suboptimal settings, while achieving similar performance.

The fitting approach based on optimal hyperparameters enables the derivation of a generalized formulation for precise predictions of achievable validation loss across different model sizes and training budgets. Our analysis shows that, with a fixed compute budget, validation loss remains relatively stable when model size is adjusted near the optimal size. This suggests that our formulation can significantly reduce inference costs in practical applications, with only a minor and predictable degradation in performance. Additionally, our results provide accurate extrapolation of the relationship between optimal model size and compute budget. In contrast, using fixed suboptimal hyperparameters results in less accurate extrapolation, as demonstrated in ~\Cref{fig:combine_Nopt}.

In summary, our contributions and findings can be summarized as follows.
\begin{itemize}
    \item We confirm the existence of scaling laws in video diffusion transformers.
    \item We establish new scaling laws for optimal hyperparameters in video diffusion transformers, enabling precise predictions of the optimal batch size and learning rate for any model size and compute budget, offering improved guidance for efficient training.
    \item Based on optimal hyperparameters, we establish a more generalized  relationship among validation loss, model size, and training budget, enabling precise guidance for trade-offs between model size and performance in practical applications.
\end{itemize}
\section{Setup}
\paragraph{Notations.}
 We will investigate scaling laws for video diffusion transformers from three critical perspectives: selecting the optimal batch size and learning rate to optimize performance, predicting the performance, and determining the optimal model/data allocation strategy to balance performance and compute budgets. To facilitate analysis in subsequent sections, we introduce the following notations:  
\begin{itemize}
    \item $L$ - the loss function for the video DiT on the validation set.
    \item $N$ - the number of model parameters, excluding all vocabulary and positional embeddings.
    \item $n_{\text{ctx}}$ - the context window length, defined as $n_{\text{ctx}} = fhw$, where $f$ is the number of frames, and $h$ and $w$ are the height and width of each frame, respectively.
    \item $C_{\text{token}} = \frac{3}{4} N \left(7 + \frac{n_{\text{ctx}}}{d}\right)$ - The compute cost per token during training, where $d$ represents the model dimension.
    \item $T$ - the number of training tokens, representing the dataset size measured in tokens.
    \item $C \approx C_{\text{token}} T$ - an estimate of the total non-embedding training compute.
\end{itemize}
A more detailed per-operation parameter and compute count is included in~\Cref{appendix:Compute}.

\paragraph{Hyperparameters.}
\cite{bi2024deepseek} demonstrate that final performance with a multi-step learning rate scheduler is largely comparable to that of a cosine scheduler. To simplify the problem and control variables, we focus our study on the stable phase of training, where we use a constant learning rate. Additionally, we find that a straightforward constant learning rate scheduler yields effective performance in predicting the optimal hyperparameters.

\paragraph{Evaluation.}
To assess model performance, we partition the original dataset to create a validation set and use validation loss as the primary evaluation metric, measuring it every $200$ training steps. We did not include external benchmarks due to challenges in consistently aligning multiple video benchmarks. Moreover, validation loss serves as an effective proxy for performance, as supported by studies like~\citep{esser2024scaling,polyak2024movie}, making it a reasonable choice for this evaluation. 

\section{Predicting Optimal Hyperparameters}
\label{sec:hyper}
Previous studies, such as ~\citep{goyal2017accurate, mccandlish2018empirical, shallue2019measuring, smith2017don}, have offered empirical insights into the batch size and learning rate relationship in large language model training. However, these investigations generally rely on heuristic methods, lacking rigorous theoretical foundations for optimal hyperparameter selection. Recent studies~\citep{yang2022tensor,blake2024u} primarily focus on hyperparameter transfer, where hyperparameters are indirectly tuned on smaller models and then transferred to full-sized models without further tuning. However, most scaling law research ignores the impact of selecting hyperparameters for different model sizes, which can significantly affect fitting precision. Specifically, these works exhibit the following limitations when transfer to video DiT:

\begin{itemize}
    \item \textbf{OpenAI's Scaling Law}~\citep{kaplan2020scaling} suggests that smaller batch sizes are more computationally efficient but require more update steps to converge. Our experiments show that in video generation, scaling laws are highly sensitive to hyperparameters
    like batch size and learning rate. As shown in~\Cref{fig:SubOptimalHp}, fixed suboptimal settings (gray points) lead to higher validation loss, while optimal settings (red points) align with the fitted loss curve, ensuring accurate predictions. Notably, these optimal hyperparameters vary with model size and training tokens. This demonstrates that there exists an optimal combination of hyperparameters that minimizes training loss and accurately captures scaling behavior. Our theoretical analyses further support the existence of these optimal hyperparameters.

    \item \textbf{Chinchilla's Scaling Law}~\citep{hoffmann2022training} establishes the relationship between the optimal model size \( N_{\text{opt}} \) and compute budget \( C \) by linking the fitted loss function to \( C \) and minimizing it under a compute constraint. However, this approach shows discrepancies compared to results obtained using IsoFLOP curves, which we attribute to imprecise loss fitting. We will provide a detailed explanation in~\Cref{sec:uncovering}.
    \item \textbf{DeepSeek's Scaling Law}~\citep{bi2024deepseek} introduces a new perspective by suggesting that, for a given compute budget, there exists an optimal combination of batch size and learning rate that maximizes performance. While this approach improves predictability in hyperparameter selection, it overlooks the influence of model size and data size, limiting its applicability in comprehensive scaling law analysis.
\end{itemize} 
These analyses underscore the importance of a refined approach to hyperparameter selection in scaling law research. To address this, we propose the precise Scaling Law for video DiT (see ~\Cref{appendix:t2i} for similar results on image DiT with 1 frame). In this section, we derive a power-law relationship linking hyperparameters to model size and data size, offering a theory-driven improvement over prior methods. This serves as the basis for our scaling law in~\Cref{sec:scaling law}.

\begin{figure*}[htbp]
    \centering
    \includegraphics[width=1.0\textwidth]{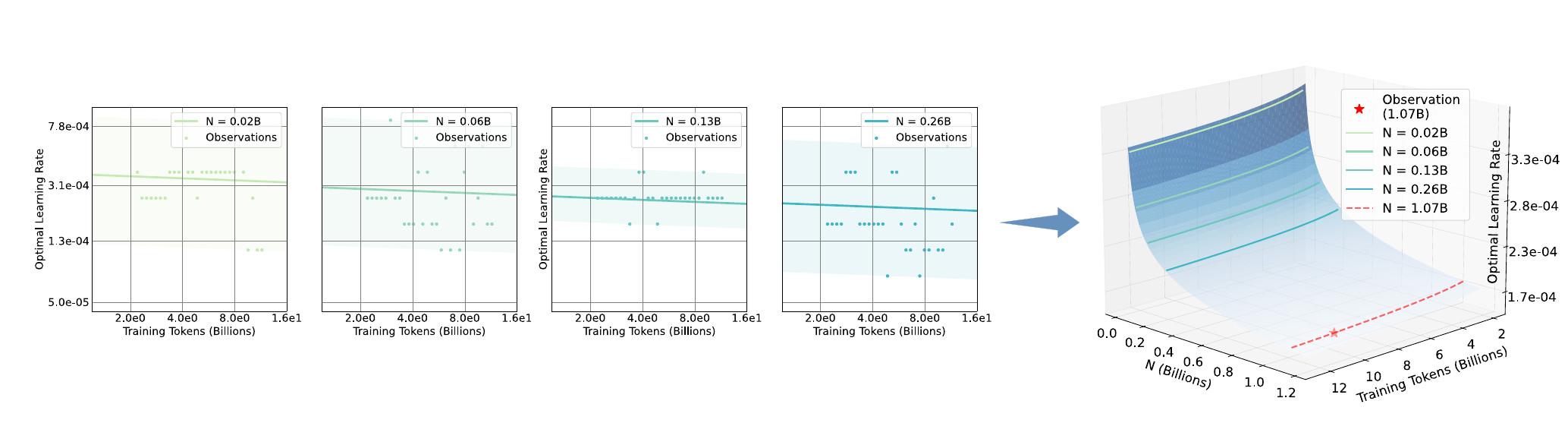}
    \vspace{-2em}
    \caption{Optimal learning rate scaling curve. \textbf{Left}: Optimal learning rate scaling curves fitted on four different model sizes (0.02B, 0.06B, 0.13B, and 0.26B parameters). ``Observations'' indicates values within 0.02\% of the minimum loss for each model size. \textbf{Right}: Extrapolated scaling curves for learning rate, predicting optimal values for a 1.07B model to achieve minimal validation loss.}
    \label{fig:optimal_lr}
    \vspace{-0.5em}
\end{figure*}

\subsection{Impact of model size on hyperparameters}
We first investigate whether model size should be considered an independent variable in the scaling laws of batch size and learning rate.

To optimize the objective function \(L(\theta)\), we need to compute its gradient \(G(\theta) = \nabla L(\theta)\). Since the objective function \( L(\theta) \) is complex, we approximate its gradient using the stochastic function. At iteration \(k\), we randomly sample a mini-batch of data \(\{\xi_k^{(b)}\}_{b=1}^B\) from the training data distribution \(\rho\). Using these samples, we compute an estimated gradient \(g_k\) as:

\begin{equation}
    g_k=\frac{1}{B} \sum_{b=1}^{B} G_{\text{est}}(\theta_k; \xi_{k}^{(b)})
    \label{eq:batch_size}
\end{equation}
where \( \xi_k^{(b)} \sim \rho \) represents a random data sampled from a distribution $\rho$ at iteration $k$ and $G_{\text{est}}(\theta_k; \xi_k^{(b)})$ is the stochastic gradient estimate for a single sample \(\xi_k^{(b)}\). 

Let $\mathcal{G}_k^{B} = \{\theta_k, \{\xi_{k-1}^{(b)}\}_{b=1}^B, \theta_{k-1}, \{\xi_{k-2}^{(b)}\}_{b=1}^B, \dots, \theta_0\}$
represent the filtration containing all historical variables at and before iteration $k$. Following \cite{mccandlish2018empirical}, we assume the estimated gradient is an unbiased estimate of the true gradient, while the variance of the estimated gradient is bounded:
% we introduce the following assumptions to facilitate the convergence analysis:
% \begin{assumption}
% \label{assume:unbias}
% Given the filtration $\mathcal{G}_k^B$, we assume that:
% \vspace{-0.5em}
% \begin{equation}
%     \mathbb{E}_{\xi_k \sim \rho}[ g_k \mid \mathcal{G}_k^B] = G(\theta_k) 
%     \label{eq:unbiased1}
% \end{equation}
% \begin{equation}
%     \mathbb{E}_{\xi_k \sim \rho}[\|g_k - G(\theta_k) \|^2 \mid \mathcal{G}_k^B] \leq \sigma_B^2 = \frac{\sigma^2}{B}
%     \label{eq:variance1}
% \end{equation}
% \end{assumption}

\begin{equation}
    \mathbb{E}_{\xi_k \sim \rho}[ g_k \mid \mathcal{G}_k^B] = G(\theta_k) 
    \label{eq:unbiased1}
\end{equation}
\begin{equation}
    \mathbb{E}_{\xi_k \sim \rho}[\|g_k - G(\theta_k) \|^2 \mid \mathcal{G}_k^B] \leq \sigma_B^2 = \frac{\sigma^2}{B}
    \label{eq:variance1}
\end{equation}

Suppose \( L(\theta_k) \) is \( L \)-smooth for any iteration k ~\cite{wang2020adapting} and \Cref{eq:unbiased1,eq:variance1} holds, we derive the convergence properties under the framework of mini-batch stochastic gradient descent (SGD) as in~\citep{mccandlish2018empirical}. For effective updates, the learning rate \( \eta \) must satisfy \( \eta \leq 1/L \).
Under these conditions, the convergence of $L(\theta_k)$ is characterized as follows (proof provided in~\Cref{appendix:proof}):
\begin{align}
    \frac{1}{K+1} \sum_{k=0}^K \mathbb{E}[\|G(\theta_k)\|^2] &\leq \frac{2 \Delta_0}{\eta (K+1)} + L \eta \sigma_B^2
    \label{eq:converge}
\end{align}
Here, \( \Delta_0 = L(\theta_0) - L^{\star} \) represents the initial gap to the optimal objective value $L^{\star}$, and \( L^{\star} \) denotes the optimal convergence point.

As the model size increases, its complexity also grows, leading to a corresponding increase in the Lipschitz constant \( L \)~\cite{herrera2020local}. To ensure that the upper bound on the right-hand side of~\Cref{eq:converge} remains manageable, both the learning rate and batch size must be adjusted with respect to $L$. Specifically, a larger $L$ necessitates a smaller learning rate $\eta$ and a larger batch size $B$ for convergence. 
This relationship illustrates how the scaling of model size impacts the optimal choice of hyperparameters. Unlike prior work~\citep{bi2024deepseek}, which treats model size implicitly, our analysis explicitly incorporates model size as an independent variable. This enables a more accurate estimation of optimal hyperparameters by addressing the interplay between model complexity, learning rate, and batch size in a principled manner.

\subsection{Proposed $B_\text{opt}(N,T)$ equation}
\begin{figure*}[htbp]
    \centering
    \includegraphics[width=1.0\textwidth]{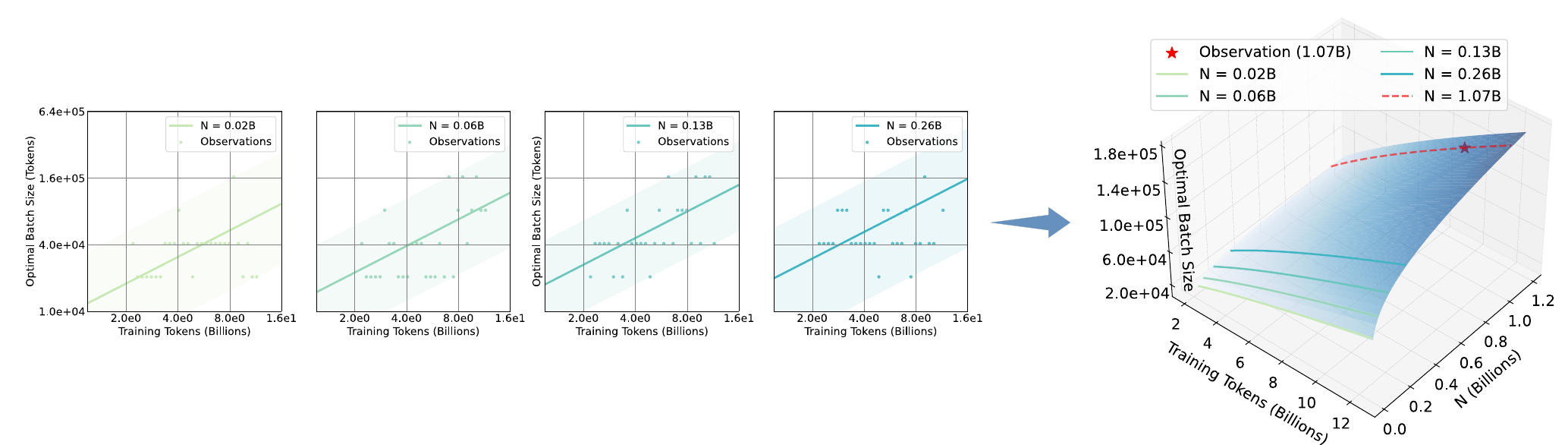}
     \vspace{-2em}
    \caption{Optimal batch size scaling curve. \textbf{Left}: Optimal batch size scaling curves fitted on four different model sizes (0.02B, 0.06B, 0.13B, and 0.26B parameters).  ``Observations" indicates values within 0.02\% of the minimum loss for each model size. \textbf{Right}: Extrapolated scaling curves for batch size, predicting optimal values for a 1.07B model to achieve minimal validation loss.}
    \label{fig:optimal_bs}
\end{figure*}
Firstly, we examine the variance of the gradient estimate in Mini-Batch SGD, formulated as,
\begin{equation}
    \mathbb{D}_{\xi_k \sim \rho}[g_k] = \frac{1}{B} \Sigma_k
    \label{eq:Gest_var}
\end{equation}
where $\Sigma_k$ is the covariance matrix of a single sample gradient, defined as:
\begin{align}
    \Sigma_k &= \mathbb{E}_{\xi_k^{(b)} \sim \rho} \left[ G_{\text{est}}(\theta_k,\xi_k^{(b)}) G_{\text{est}}(\theta_k,\xi_k^{(b)})^\top \right] \nonumber \\
    &- G(\theta_k) G(\theta_k)^T 
    \label{eq:Sigma}
\end{align}

\paragraph{Stepwise loss optimization.}
When the model is updated with the learning rate \(\eta\), the stepwise loss at iteration k is given as follows, with detailed proofs provided in~\Cref{appendix:proof}:

\begin{small} 
\begin{align}
    \Delta L_k &= \mathbb{E} \left[ L(\theta_{k-1} - \eta g_k) \right] - L(\theta_{k-1}) \nonumber \\
    & \approx - \eta \|G(\theta_k)\|^2 + \frac{1}{2} \eta^2 \left( G(\theta_k)^T H_k G(\theta_k) + \frac{\text{tr}(H_k \Sigma_k)}{B} \right)
    \label{eq:lossstep1}
\end{align}
\end{small}
The optimal learning rate for minimizing \Cref{eq:lossstep1} is given by:
\begin{small}
\begin{equation}
    \eta_{\text{opt}}(B) = \underset{\eta}{\operatorname{argmin}} \Delta L_k = \frac{\|G(\theta_k)\|^2 }{G(\theta_k)^T H_k G(\theta_k) + \frac{\text{tr}(H_k \Sigma_k)}{B}} 
    \label{eq:lropt}
\end{equation}
\end{small}
The corresponding maximum change in loss, $\Delta L_k^\ast$, is expressed as:
\begin{equation}
    \Delta L_k^\ast = -\frac{\|G(\theta_k)\|^4 }{2 \left( G(\theta_k)^T H_k G(\theta_k) + \frac{\text{tr}(H_k \Sigma_k)}{B} \right)}
    \label{eq:max_update}
\end{equation}
For simplicity, we refer to \( \frac{\text{tr}(H_k \Sigma_k)}{B} \) as the \textit{gradient noise}, which diminishes with larger batch sizes.

\paragraph{Trade-off.}
Given a compute budget \( C \) and model size \( N \), the total number of training tokens is defined as \( T = BS = \frac{C}{C_{\text{token}}} \). Adjusting the batch size \( B \) requires balancing \( S \) and the gradient noise at each step. As \( B \) increases, the variance of the gradient estimate decreases, improving the gain of each step. However, since \( S = \frac{C}{B C_{\text{token}}} \), the total number of update steps \( S \) decreases, which limits the cumulative gains over the total steps. Therefore, \textit{an optimal batch size \( B \) is a balance between per-step improvement and the number of update steps \( S \) to achieve the maximum total updates.}

\paragraph{Scaling law for optimal batch size.}
Based on the aforementioned analyses, we posit that under a fixed compute budget, there exists an optimal batch size \( B_{\text{opt}} \) that achieves the best trade-off between gradient noise and the number of update steps, with the selection of the optimal batch size primarily influenced by the model size \( N \) and the total training tokens \( T \). Referring to \cite{bi2024deepseek}, we provide a concise formulation:
\begin{equation}
    B_{\text{opt}} = \alpha_B T^{\beta_B} N^{\gamma_B}
    \label{eq:Bopt1}
\end{equation}

\subsection{Proposed $\eta_\text{opt}(N,T)$ equation}
To determine an effective learning rate, we first consider the constraints that ensure each update step reduces the loss. Specifically, the expected change in loss at the $k$-th step, $\Delta L_k = \mathbb{E}[L(\theta_k - \eta g_k)] - L(\theta_{k-1})$, must satisfy $\Delta L_k<0$. From this, we derive the following condition for the learning rate:
\begin{equation}
    \eta(B) < \frac{2 \|G(\theta_k)\|^2 }{G(\theta_k)^T H_k G(\theta_k) + \frac{\text{tr}(H_k \Sigma_k)}{B}}
\label{eq:lr_restrict}
\end{equation}
As training progresses, the gradients become smaller, leading to tighter bounds on the learning rate to ensure effective updates.
\paragraph{Diminishing returns over training steps.}
Consider a training process with \( S \) steps, where the sequences of upper bounds on effective learning rates and optimal learning rates are denoted as:
\begin{equation}
    \eta_{\text{bound}} = [\eta, \eta_2, \ldots, \eta_S],\quad
    \eta_{\text{opt}} = [\tilde{\eta}_1, \tilde{\eta}_2, \ldots, \tilde{\eta}_S]
    \label{eq:eta_bound_opt}
\end{equation}
Ignoring sampling noise, both $\eta_{\text{bound}}$ and $\eta_{\text{opt}}$ tend to decrease over the course of training, as the model approaches convergence and the gradients become smaller. Concurrently, the maximum per-step gain $-\Delta L_k^{*}$ also tends to decrease as the model converges.
\paragraph{Trade-off.}
With a constant learning rate \(\eta^*\), when \(\eta^*\) closely matches the values in \(\eta_{\text{opt}}\), it maximizes the per-step gains. However, as shown in~\Cref{eq:max_update}, these gains diminish as the model converges, and \(\eta^*\) may exceed the values in \(\eta_{\text{bound}}\), leading to ineffective updates. On the other hand, a smaller \(\eta^*\), while ensuring more effective steps, results in slower initial progress and reduced overall efficiency. Therefore, \textit{an optimal learning rate \( \eta_{\text{opt}} \) balances per-step gains with the number of effective update steps to achieve maximum overall improvement.}

\paragraph{Scaling law for optimal learning rate.}
Similar to the optimal batch size, the optimal learning rate aims to minimize \( L(\theta) \) under a fixed number of training tokens \( T \) by balancing the number of effective steps with the gain per step. Thus, we express the optimal learning rate as \(\eta_{\text{opt}} \propto (\frac{T}{B_{\text{opt}}})^{k_1} N^{k_2}\). Based on \Cref{eq:Bopt1}, we derive:

\begin{equation}
    \eta_{\text{opt}} = \alpha_\eta T^{\beta_\eta} N^{\gamma_\eta}
    \label{eq:lr_opt1}
\end{equation}

\subsection{Results of optimal hyperparameters}
\begin{figure*}[htbp]
    \centering
    \includegraphics[width=0.9\textwidth]{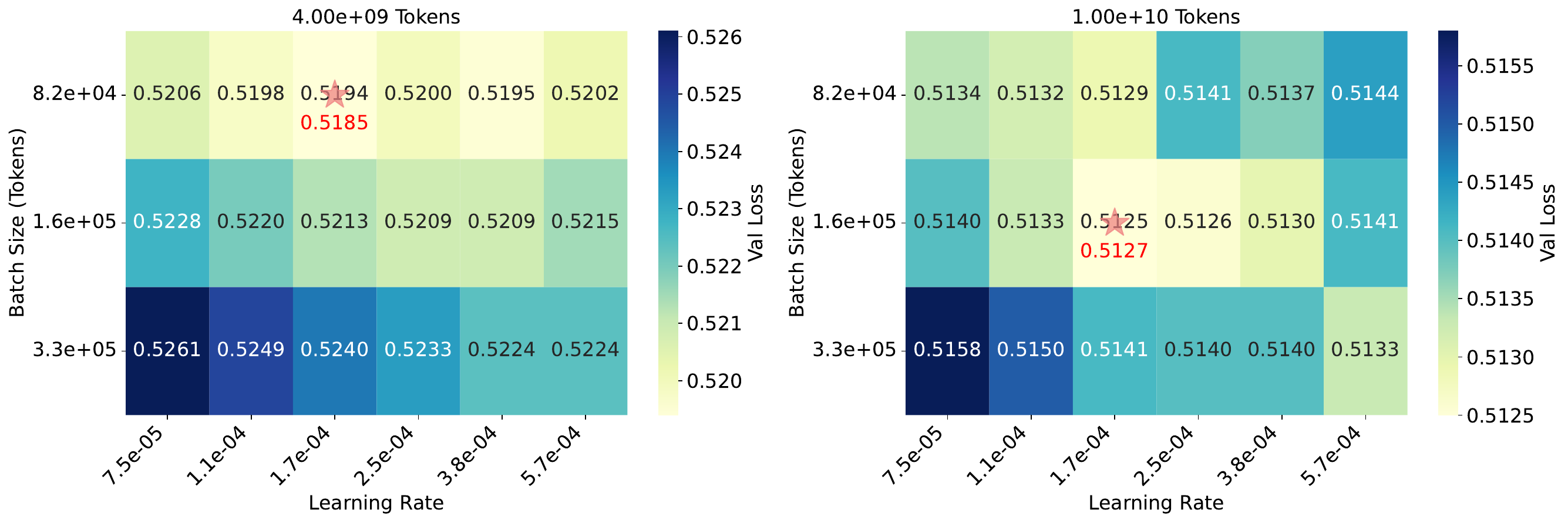}
    \caption{Predictions of optimal hyperparameters on 1.07B model size with 4B and 10B training tokens. The red pentagrams indicate the predicted optimal batch size and learning rate, along with their predicted validation loss.}
    \label{fig:GridVal}
\end{figure*}
In our experiments, we obtained fitting results across various hyperparameter combinations for model sizes of 0.017B, 0.057B, 0.13B, and 0.26B parameters, using training tokens from 2B to 12B. Data points for fitting included all configurations with loss values within \(0.02\%\) of the minimum, as shown in \Cref{fig:optimal_bs,fig:optimal_lr}. The final fitted parameter values are summarized in \Cref{tab:fitting_results_bopt,tab:fitting_results_etaopt}.

The results indicate that larger model sizes benefit from larger batch sizes and lower learning rates. Additionally, as the number of training tokens increases, it is possible to increase the batch size, accepting a slight reduction in update steps to reduce gradient noise and enhance per-step gains. For learning rate, a moderate decrease can be applied, which may slightly reduce initial gains but ultimately results in more effective update steps. These trends are consistent with \cite{bi2024deepseek}, and our approach provides more accurate and reliable predictions across any model sizes and training tokens.
\begin{table}[h!]
    \centering
    \renewcommand{\arraystretch}{1.2} % Adjust row height
    \setlength{\tabcolsep}{6pt} % Adjust column spacing
    \begin{tabular}{|c|c|c|c|}
        \hline
        \textbf{Parameter} & \( \alpha_B \) & \( \beta_B \) & \( \gamma_B \) \\
        \hline
        \textbf{Value} & \( 2.1797 \times 10^4 \) & \( 0.8080 \) & \( 0.1906 \) \\
        \hline
    \end{tabular}
    \caption{Fitting Results for \( B_{\text{opt}} \)}
    \label{tab:fitting_results_bopt}
\end{table}

To validate the accuracy of our fitting results, we expanded the model size fourfold to 1.07B with 4B and 10B training tokens. We conducted a grid search with various configurations of batch size and learning rate, confirming that the predicted optimal batch size and learning rate indeed minimized the validation loss, as shown in~\Cref{fig:GridVal}. 

\begin{table}[h!]
    \centering
    \renewcommand{\arraystretch}{1.2} % Adjust row height
    \setlength{\tabcolsep}{6pt} % Adjust column spacing
    \begin{tabular}{|c|c|c|c|}
        \hline
        \textbf{Parameter} & \( \alpha_\eta \) & \( \beta_\eta \) & \( \gamma_\eta \) \\
        \hline
        \textbf{Value} & \( 0.0002 \) & \( -0.0453 \) & \( -0.1619 \) \\
        \hline
    \end{tabular}
    \caption{Fitting Results for \( \eta_{\text{opt}} \)}
    \label{tab:fitting_results_etaopt}
\end{table}
 
\section{Scaling Law in Video DiT with Optimal Hyperparameters}
\label{sec:scaling law}
In this section, we will predict the optimal model size under a fixed compute budget, utilizing optimal hyperparameters. Concurrently, we will fit the validation loss for any model size and training tokens under optimal hyperparameters. This not only allows us to accurately predict the validation loss at the optimal model size but also enables us to predict validation loss for any fixed suboptimal model size and compute budget, which significantly enhances the practical value of the fitting results as we always forgo the optimal model size due to hardware constraints and inference costs.

\subsection{Scaling law for optimal model size}
\begin{figure*}[thb]
    \centering
    % 第一张子图
    \begin{subfigure}[b]{0.31\textwidth}
        \centering
        \includegraphics[width=\textwidth]{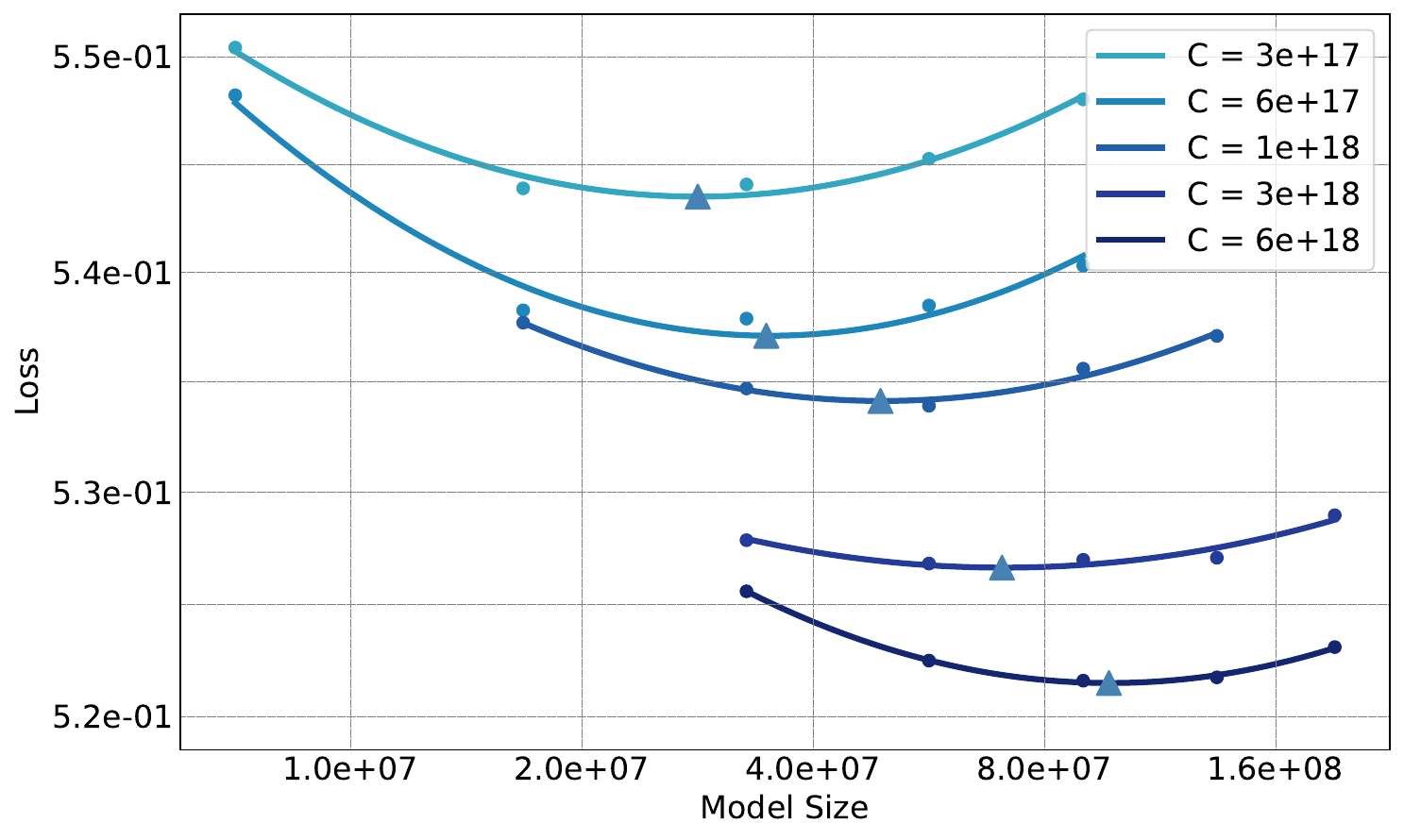}
        \caption{Optimal hyperparameters.}
        \label{fig:Nopt_exp1}
    \end{subfigure}
    \hfill
    \begin{subfigure}[b]{0.31\textwidth}
        \centering
        \includegraphics[width=\textwidth]{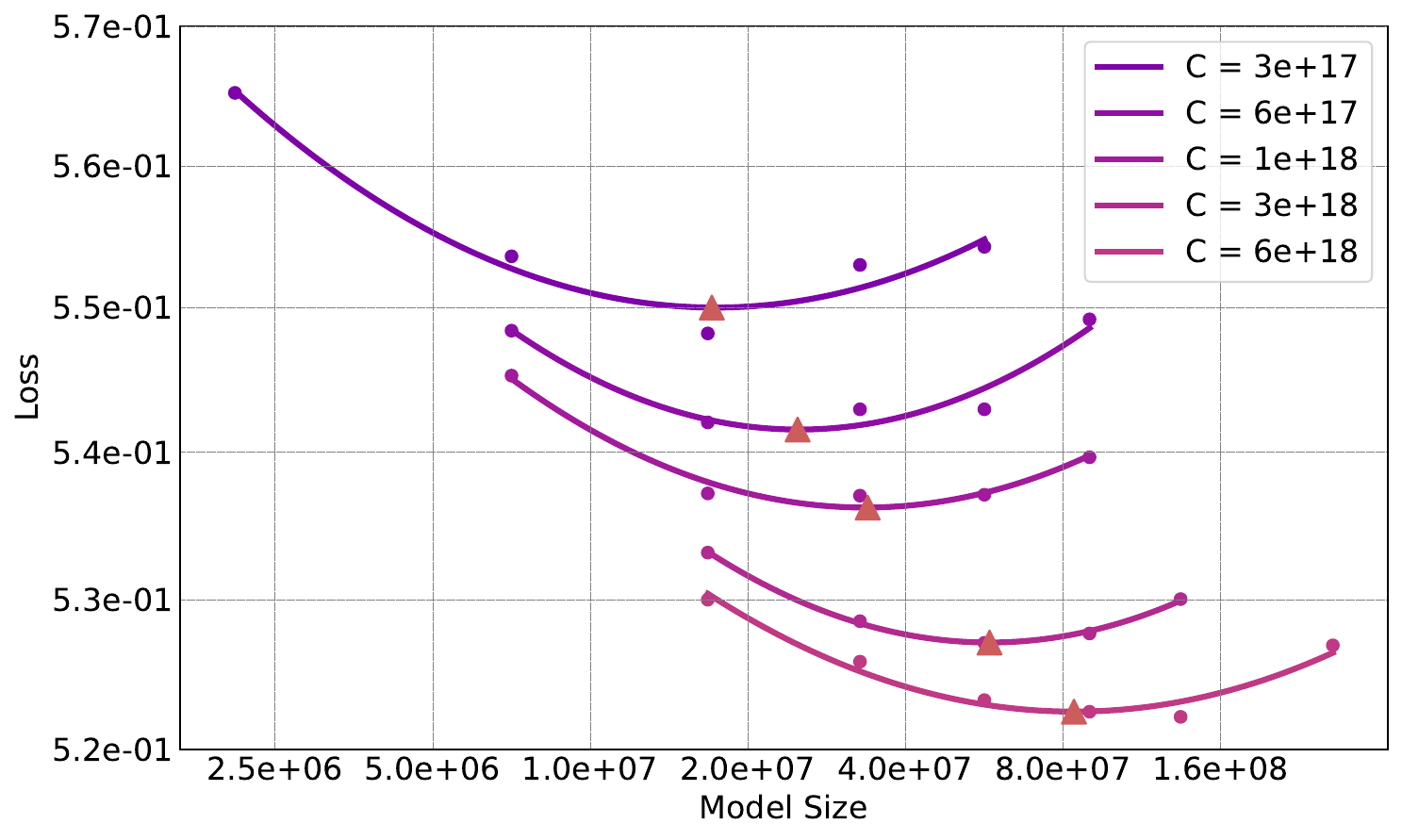}
        \caption{Fixed suboptimal hyperparameters.}
        \label{fig:FixedHp_Nopt_exp}
    \end{subfigure}
    \hfill
    \begin{subfigure}[b]{0.3\textwidth}
        \centering
        \includegraphics[width=\textwidth]{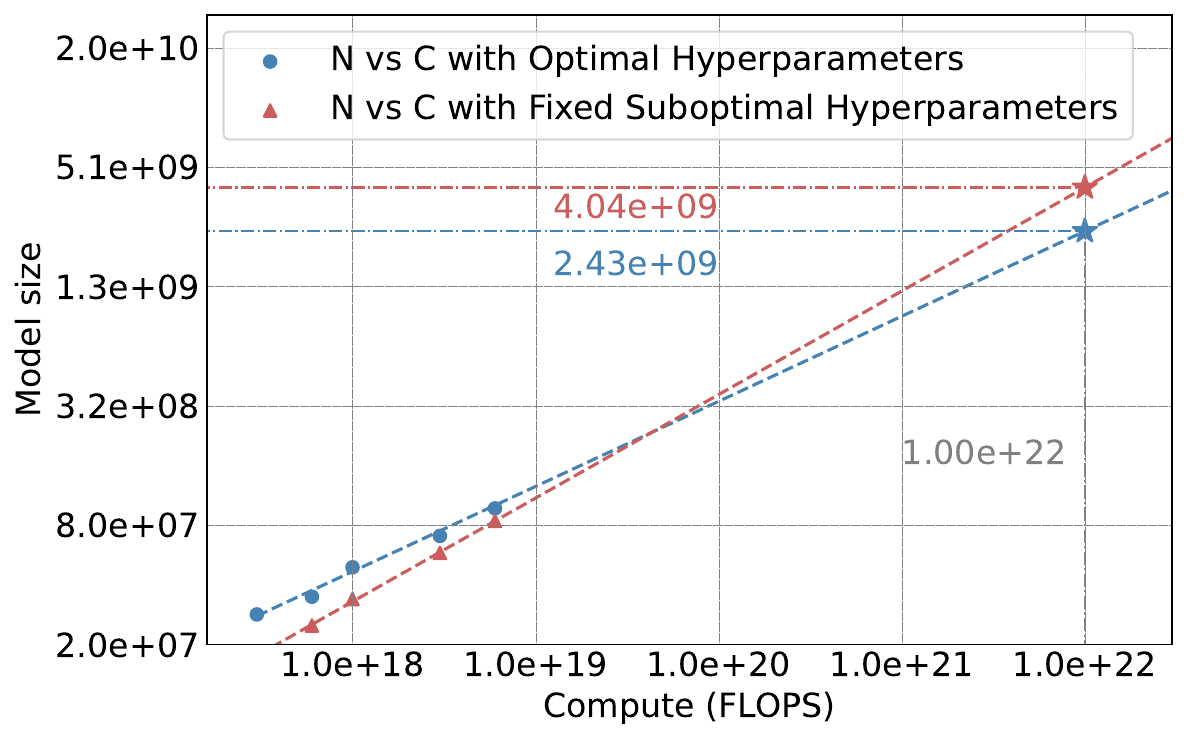}
        \caption{Empirical loss v.s. \( N \).}
        \label{fig:Nopt_fixedNopt}
    \end{subfigure}
    \vspace{-0.5em}
    \caption{Comparison of empirical loss v.s. \( N \) under optimal and fixed suboptimal hyperparameters. \textbf{Left}: Empirical loss as a function of model size $N$ for various compute budgets $C$ under optimal hyperparameters, with a parabolic fit to identify minimum loss points.
    \textbf{Middle}: Results under fixed suboptimal hyperparameters. \textbf{Right}: We contrasted the fitting results obtained with optimal and fixed suboptimal hyperparameters. The use of suboptimal hyperparameters leads to an overestimation of the optimal model size, resulting in wasted computational resources and degraded final model performance.}
    \label{fig:precise_optimal_model_size}
\end{figure*}

In this section, we explore how to optimally allocate model size $N$ and training tokens $T$ within a given compute budget $C$ to maximize model performance. Building on the approach proposed in~\citep{bi2024deepseek}, we leverage the optimal hyperparameters defined in~\Cref{eq:lr_opt1,eq:Bopt1} for our experiments. For each compute budget \([3e17, 6e17, 1e18, 3e18, 6e18]\), we conducted five experiments to ensure reliable results. As shown in~\Cref{fig:Nopt_exp1}, we fit a parabola to each compute budget’s results to identify the minimum loss point, which corresponds to \textbf{\textit{empirical optimal model size}} under that specific budget.
\begin{equation}
    N_{\text{opt}} = 1.5787 \cdot C^{0.4146}
    \label{eq:Nopt_exp_1} 
\end{equation}
We will validate the accuracy of our results by extrapolating the validation loss in the subsequent sections.
\begin{figure*}
    \centering
    \begin{subfigure}[t]{0.48\textwidth}
        \centering
        \includegraphics[scale=0.3]{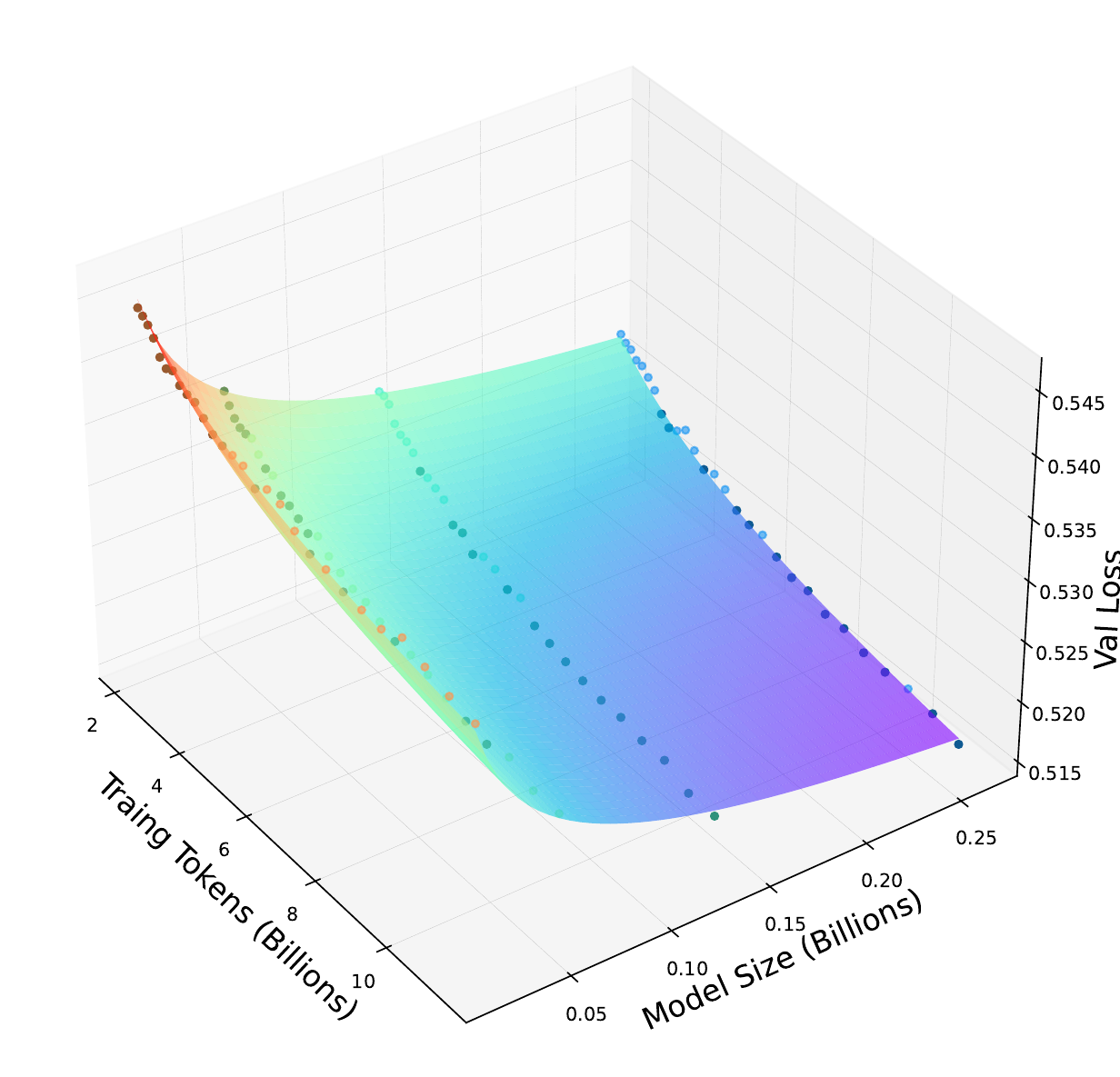}
        \caption{Performance scaling curve fitted on four small model scales.}
        \label{fig:Loss_3D_1}
    \end{subfigure}
    \hfill
    \begin{subfigure}[t]{0.48\textwidth}
        \centering
        \includegraphics[width=\textwidth]{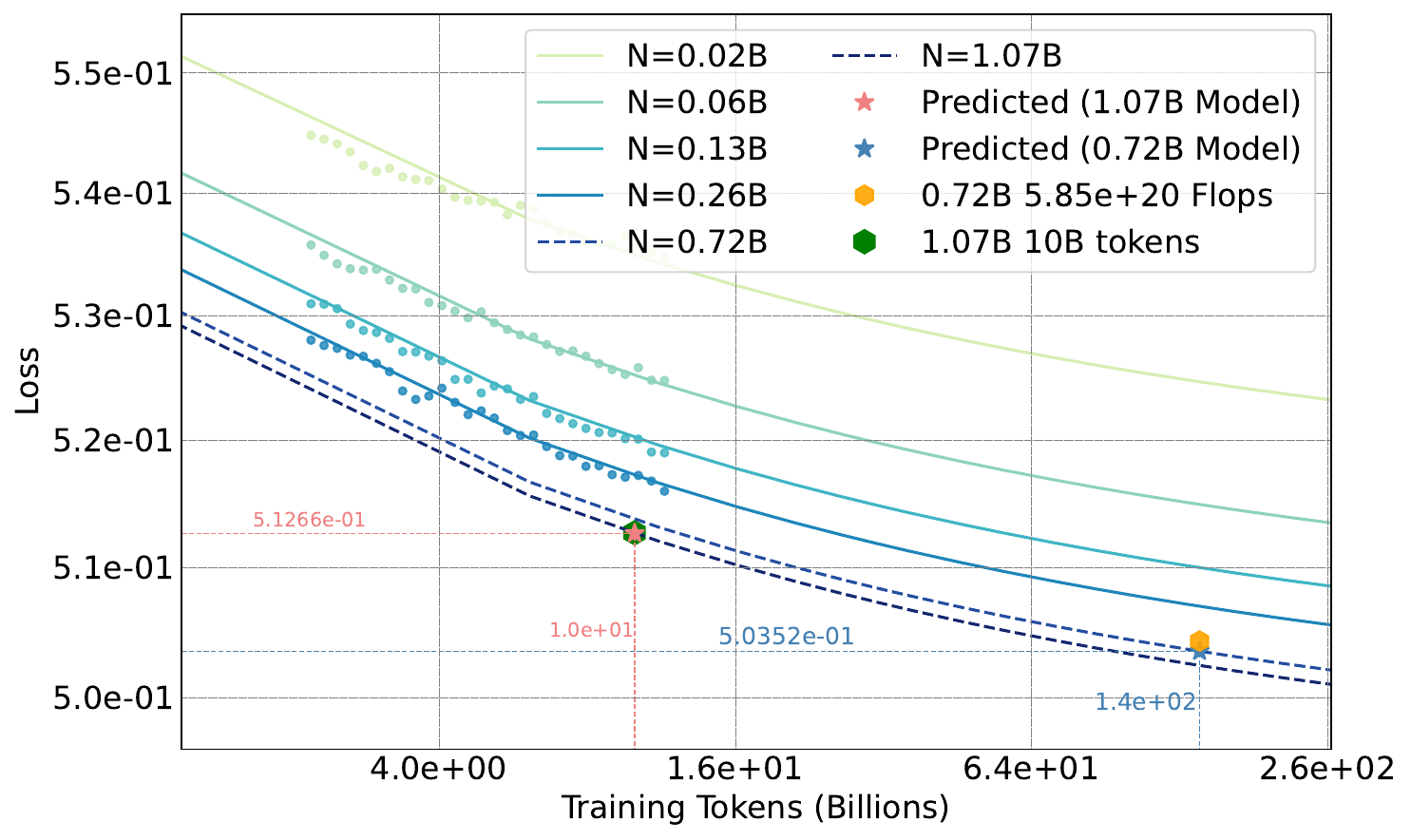}
        \caption{Performance scaling curve extrapolated to larger models.}
        \label{fig:Loss_1}
    \end{subfigure}
    \caption{Loss scaling with optimal hyperparameters across varying model and compute scales. \textbf{Left}: Fitted loss curves under optimal hyperparameters across four smaller models, each trained with varying numbers of tokens. \textbf{Right}: Extrapolated loss curves extended to larger model sizes and compute budgets, offering predictions for any model size and traing tokens. The red pentagram indicates the projected loss for a 1.07B model with 10B training tokens, while the blue pentagram marks the expected loss for a 0.72B model under a compute budget of \(5.85 \times 10^{20}\). Experimental results are shown as green and orange hexagons, validating the extrapolation accuracy.}
    \label{fig:Loss_combine1}
\end{figure*}
\subsection{Scaling law for performance}
\cite{henighan2020scaling,hoffmann2022training} modeled validation loss as a function of model size $N$ and training steps $S$ or data size $D$, but did not detail hyperparameter settings, leaving it unclear if models achieved optimal performance across different compute budgets. \cite{bi2024deepseek} introduced a simplified power-law relationship between loss $L$ and compute budget $C$, assuming optimal hyperparameters and model size. However, in practice, the optimal model size is often not used due to hardware limitations or the need to balance performance with model size. Thus, fitting the loss only at the optimal model size is insufficient for practical needs.

To address these limitations, we propose a more general and accurate performance prediction method, capable of estimating loss across any model size and compute budget while using optimal hyperparameters. Similar to \cite{hoffmann2022training}, the following fitting formula is designed to adhere to two guiding principles:

\begin{equation}
    L(T, N) = \left(\frac{T_c}{T}\right)^{\alpha_T} + \left(\frac{N_c}{N}\right)^{\alpha_N} + L_\infty
    \label{eq:loss1}
\end{equation}

\begin{itemize}
    \item As \( N \to \infty \), the minimum achievable loss depends on the training data entropy and noise. Similarly, as \( T \to \infty \), it depends on model size.
    \item With infinite compute, as both \( N \) and \( T \) approach infinity, the loss of video diffusion transformers asymptotically approaches the training data entropy \( L_\infty \).
\end{itemize}

Our fitting results are displayed in~\Cref{tab:fitting_loss1}. As shown in~\Cref{fig:Loss_3D_1}, loss decreases with increasing model size and training tokens. To validate our fitting accuracy, we tested a 1.07B model on 10B training tokens and a 0.72B model on 140B training tokens. As shown in~\Cref{fig:Loss_1}, the deviations between our predicted and experimental results were $0.03\%$ and $0.15\%$, respectively, confirming the robustness of our approach.

\begin{table}[h!]
    \centering
    \renewcommand{\arraystretch}{1.2}  % Adjust row height
    \setlength{\tabcolsep}{4pt}  % Adjust column spacing
    \begin{tabular}{|c|c|c|c|c|c|}
        \hline
        \textbf{Parameter} & \( T_{\text{c}} \) & \( \alpha_T \) & \( N_{\text{c}} \) & \( \alpha_N \) & \( L_{\infty} \) \\
        \hline
        \textbf{Value} & $0.0373$ & $0.2917$ & $0.0082$ & $0.3188$ & $0.4856$ \\
        \hline
    \end{tabular}
    \caption{Fitting Results for \(L(T, N)\)}
    \label{tab:fitting_loss1}
\end{table}

\begin{figure*}[htbp]
    \centering
    \includegraphics[width=\linewidth]{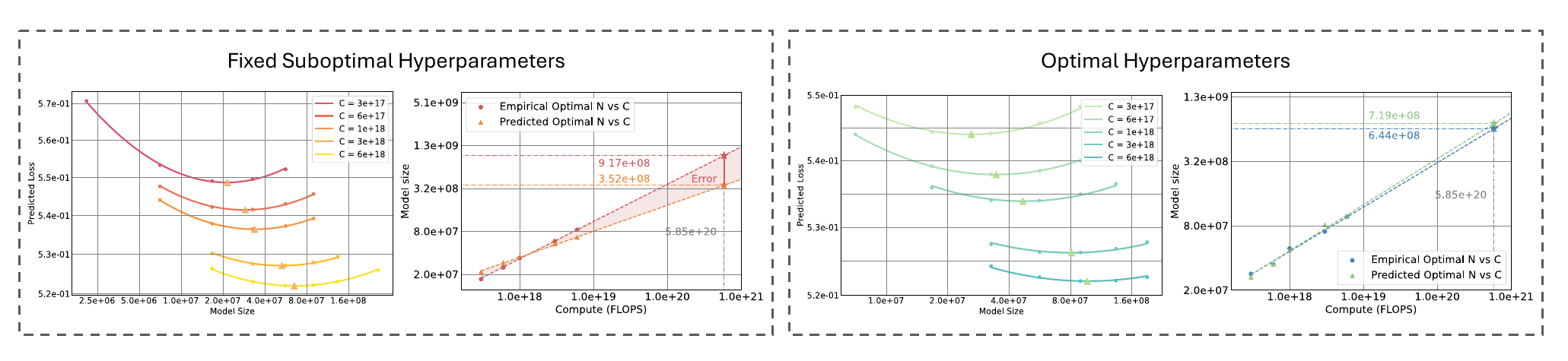}
    \caption{Comparison of empirical and predicted optimal model size scaling law under optimal and fixed suboptimal hyperparameters. \textbf{Left}: Using fixed suboptimal hyperparameters, we calculate the optimal model size from IsoFLOP profiles (empirical optimal model size) and loss function $L(N,T)$ (predicted optimal model size), obtaining the red and yellow lines. The yellow line shows a slope deviation of $30.26\%$ compared to the red line. This discrepancy arises from the insufficient accuracy of the $L(N,T)$. \textbf{Right}: When we conducted the same experiments with optimal hyperparameters, the slope deviation between the two results was reduced to $3.57\%$. }
    \label{fig:combine_Nopt}
\end{figure*}

\subsection{Uncovering the advantages}
\label{sec:uncovering}
In this section, we validate the advantages of our scaling laws under optimal hyperparameters. We demonstrate that our scaling laws precisely predict the optimal model size and validation loss by comparing results under optimal and fixed suboptimal hyperparameters. The results highlight that optimal hyperparameters significantly improve parameter efficiency and reduce prediction error. These findings establish the importance of using optimal hyperparameters in accurately modeling the relationships between model size, compute budget, and validation loss.
\subsubsection{More parameter-efficient model sizes}
We first validate our fitting results by extrapolating the compute budget to $5.85 \times 10^{20}$. Under this condition, the optimal model size predicted by~\Cref{eq:Nopt_exp_1} is 0.64B. Due to discrete layer selection, we conducted an experimental validation using a $14$-layer model with 719.3M parameters. According to equations~\Cref{eq:Bopt1,eq:lr_opt1}, we set the batch size to $832$ and the learning rate to $1.6 \times 10^{-4}$. The resulting loss closely aligned with the prediction from~\Cref{eq:loss1} (see~\Cref{fig:Loss_1}), further validating our conclusions.

To validate the optimal hyperparameters yield more precise fitting results, we conducted experiments with fixed suboptimal hyperparameters to fit the relationship between optimal model size and compute budget (detailed in \Cref{appendix:experiment}). As shown in~\Cref{fig:precise_optimal_model_size}, using fixed suboptimal hyperparameters overestimates the optimal model size, and this discrepancy grows with increasing compute budgets. At a compute budget of $1 \times 10^{10}$ TFlops, the scaling law based on optimal hyperparameters predicts a model that saves $39.9\%$ of the parameters while achieving a similar validation loss. This occurs because, when computational resources are abundant relative to data, scaling the model size near the optimal size has a minimal impact on validation loss. \textbf{\textit{Therefore, our scaling law precisely predict the optimal model size under optimal hyperparameters.}}

\subsubsection{More precise validation loss predictions}
To evaluate the precision of our scaling laws, we conducted experiments with both optimal and fixed suboptimal hyperparameters to fit the validation loss \( L(N, T) \). 

With fixed suboptimal hyperparameters, the mean squared error (MSE) between the observed validation loss and the fitted curve is \( 4.31 \times 10^{-7} \). Under optimal hyperparameters, however, the MSE decreases significantly to \( 2.35 \times 10^{-7} \), representing a \( 45.5\% \) reduction. This demonstrates the superior predictive capability of our scaling laws when using optimal hyperparameters.

Additionally, we validate the accuracy of our predictions by fitting the predicted optimal model size based on the loss function derived under both optimal and fixed suboptimal hyperparameters (~\Cref{tab:fitting_loss1,tab:fitting_loss_ap1}).
We then compare the predicted optimal model size with empirical optimal model size to assess the accuracy of the loss function. Specifically, we introduce a constraint incorporating compute budget \( C=C_{token}T \) into (~\Cref{eq:loss1}):
\begin{equation}
    \frac{C_{\text{token}}}{N} = \frac{3}{4} \left(7 + \frac{n_{ctx}}{d}\right)
    \label{eq:compute}
\end{equation}
then we have:
\begin{equation}
    L(N) = \frac{T_c \cdot 3N \left(7 + \frac{n_{ctx}}{d}\right)}{4C} + \left(\frac{N_c}{N}\right)^{\alpha_N} + L_\infty
    \label{eq:compute_relation}
\end{equation}
For each fixed \( C \), we identify the \( N \) that minimizes validation loss, known as the \textbf{\textit{predicted optimal model size}}. Adjusting \( N \) by scaling layers introduces discrete parameter changes. We approximate the optimal \( N \) by finding the minimum of a fitted quadratic function, as suggested by \citep{bi2024deepseek}. Using \Cref{eq:loss1}, we model the relationship between \( N_{\text{opt}} \) and \( C \) in a power-law form under optimal hyperparameters. 

Under optimal hyperparameters, the derived expression for the predicted optimal model size is:
\begin{equation} 
    \hat{N}_{\text{opt}} = 0.8705 \cdot C^{0.4294} 
    \label{eq:Nopt_predict1} 
\end{equation}
The corresponding results under fixed suboptimal hyperparameters are provided in ~\Cref{appendix:experiment}. 
We focus on the deviation in the exponent of 
$C$ between the predicted optimal model size and the empirical optimal model size relationships. As depicted in \Cref{fig:combine_Nopt}, under optimal hyperparameters, the deviation results in an absolute error of $0.0148$, corresponding to a $3.57\%$ slope deviation. In contrast, as shown in~\Cref{eq:Nopt_exp,eq:Nopt_predict}, the deviation under fixed suboptimal hyperparameters is significantly larger, with an absolute error of $0.1581$, and $30.26\%$ difference. This stark difference underscores the importance of using optimal hyperparameters to improve the precision of predicted relationships. \textbf{\textit{Therefore, our scaling laws precisely predict the validation loss under optimal hyperparameters.}}

% Finally, based on \Cref{eq:compute}, we derive the relationship between the optimal training tokens \( T_{\text{opt}} \) and the compute budget \( C \) as follows:
% \begin{equation}
%     \hat{T}_{\text{opt}} = \frac{4}{3\left(7 + \frac{n_{ctx}}{d}\right)} \cdot C^{0.5706}
%     \label{eq:Topt_predict}
% \end{equation}
\section{Related Work}

\paragraph{Text-to-Video Generation Models.}
Text-to-Video (T2V) generation has evolved from early GAN-based and VAE-based methods to modern architectures like diffusion models, DiT-based frameworks, and auto-regressive models. Early GAN approaches~\citep{vondrick2016generating, saito2017temporal, tulyakov2018mocogan, clark2019adversarial, yu2022generating} struggled with temporal coherence, causing frame inconsistencies. Video diffusion models (VDMs) adapted text-to-image diffusion techniques to enhance frame continuity. DiT-based architectures~\citep{peebles2023dit, lu2023vdt, ma2024latte, gao2024lumina} introduced spatio-temporal attention mechanisms, improving the capture of complex video dynamics and frame coherence~\citep{he2022latent, blattmann2023align, chen2023seine, girdhar2023emuvideo}. Auto-regressive models~\citep{yan2021videogpt, hong2022cogvideo, villegas2022phenaki, kondratyuk2023videopoet, xie2024show, liu2024mardini} use token-based methods to capture temporal dependencies, excelling in long-form video generation~\citep{yin2023nuwa, wang2023genlvideo, zhao2024moviedreamer, henschel2024streamingt2v, tan2024videoinfinity, zhou2024storydiffusion} and video-to-video translation~\citep{yang2023rerender, yatim2024smm, hu2024depthcrafter}.

\paragraph{Scaling Laws.}
Scaling laws are crucial for understanding large neural networks' performance as the increase in model and data size. While well-established for LLMs~\citep{hestness2017deep,henighan2020scalinglawsautoregressivegenerative,hoffmann2022training,kaplan2020scaling}, scaling laws in the generative domain remains insufficient. In image generation, \citep{mei2024bigger, li2024scalability} are mostly empirical, with precise scaling relationships still unclear, limiting our understanding of compute budgets, model size, dataset size, and performance. A concurrent work~\citep{liang2024scaling} attempt to establish precise scaling laws but was limited by smaller batch sizes and impractical hyperparameter configurations, reducing its applicability.  Additionally, limited research has explored scaling laws in the video domain. The temporal complexity in video data suggests that scaling behaviors in video models may differ from other modalities, necessitating further investigation into video-specfic scaling laws.
\section{Conclusion, Limitation and Future Work}
In this work, we draw several key conclusions. First, we confirmed that video DiT models exhibit scaling laws, providing foundational theoretical support for the scalability of these models and establishing a basis for future applications with larger datasets and model size. Second, we establish scaling laws for optimal hyperparameters in video diffusion transformers that can predict the optimal hyperparameters for any model size and compute budget. Third, we provide precise predictions for the optimal model size and validation loss under optimal hyperparameters.

Despite these contributions, there are limitations that suggest future research directions. 
First, we only fitted scaling laws with a constant learning rate. Investigating learning rate decay could yield different results but would require more computational resources. 
Second, our evaluation focused on validation loss due to the lack of a standard metric for video generation quality. Developing robust evaluation metrics for video quality is an important area for future research to better understand model performance in real-world applications.
Third, our study on low-resolution, smaller models may not directly apply to higher-resolution, larger models, requiring new fitting. Additionally, we did not explore how video resolution and frame rate impact the scaling law. Despite higher computational demands, future research will incorporate these factors for deeper insights.
{
    \small
    \bibliographystyle{ieeenat_fullname}
    \bibliography{main}
}
\newpage
\clearpage
\appendix
\setcounter{page}{1}
\maketitlesupplementary

\section{Experimental Settings and Main Results}

\begin{figure*}[ht]
    \centering
    \begin{subfigure}[t]{0.48\textwidth}
        \centering
        \includegraphics[scale=0.3]{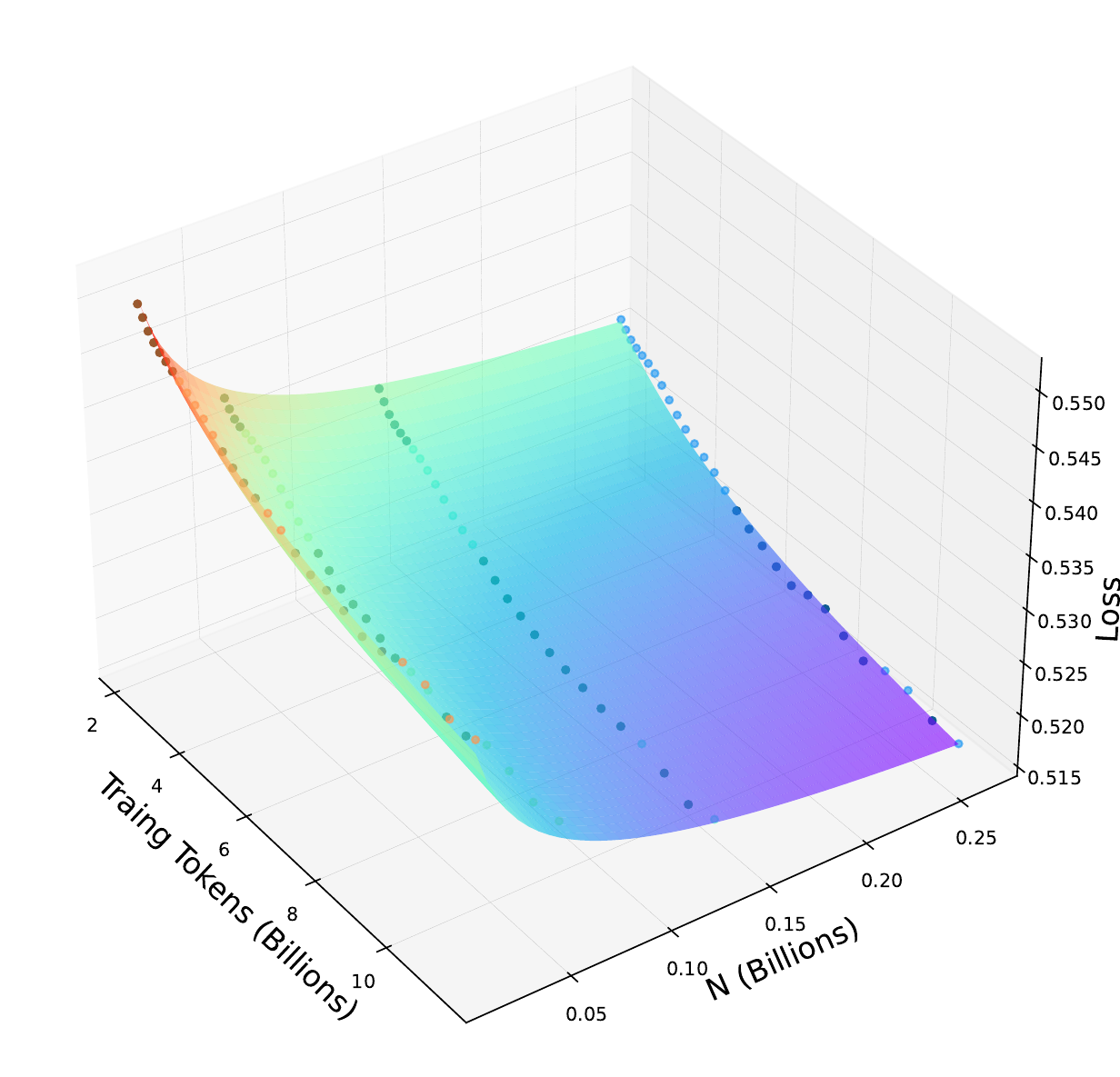}
        \caption{Performance scaling curve fitted on four small model scales.}
        \label{fig:FixedHp_Loss_3D}
    \end{subfigure}
    \hfill
    \begin{subfigure}[t]{0.48\textwidth}
        \centering
        \includegraphics[width=\textwidth]{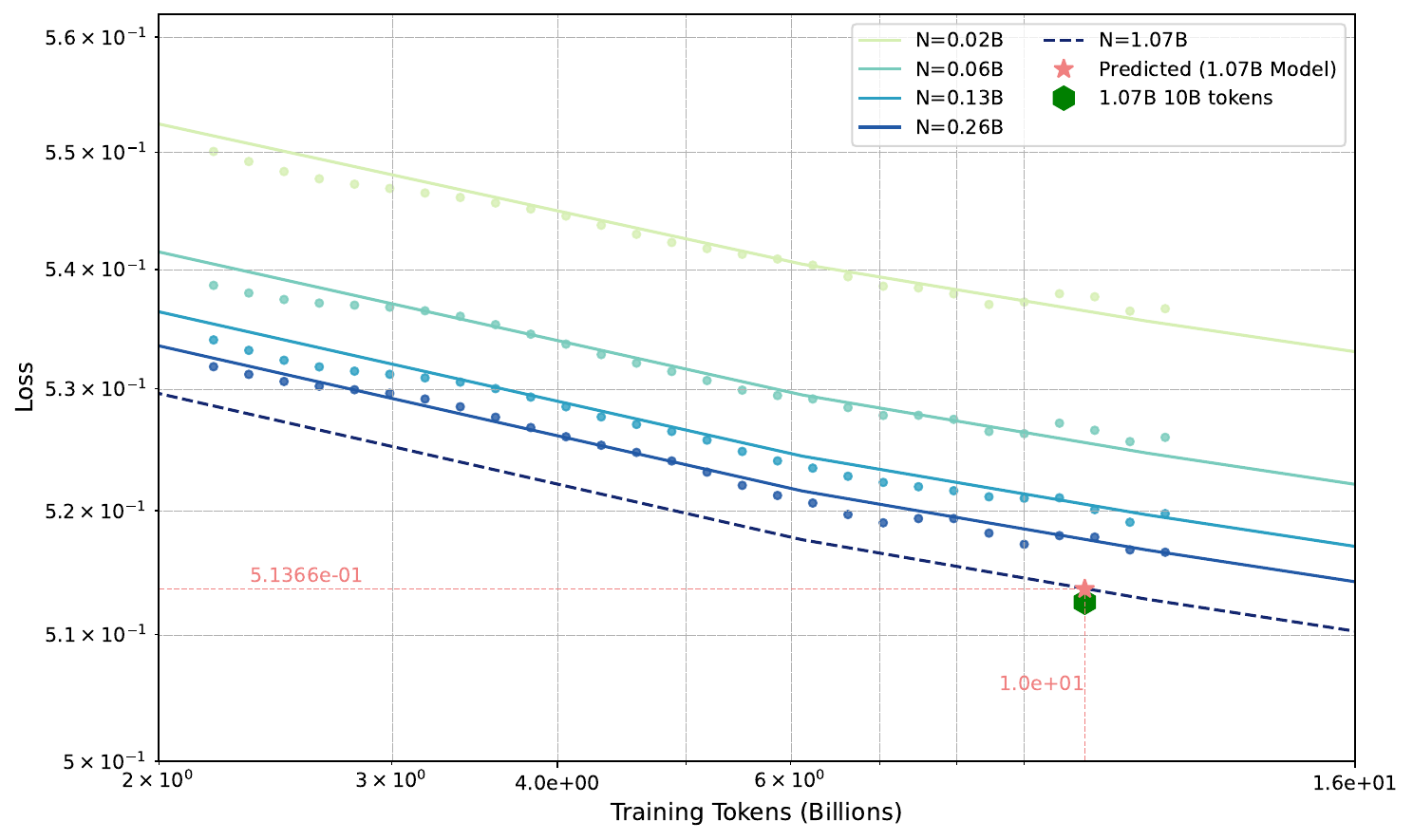}
        \caption{Performance scaling curve extrapolated to larger models.}
        \label{fig:fixed-Loss}
    \end{subfigure}
    \caption{Loss scaling with fixed suboptimal hyperparameters across varying model and compute scales. \textbf{Left}: Fitted loss curves under fixed suboptimal hyperparameters across four smaller models, each trained with varying numbers of tokens. \textbf{Right}: Extrapolated loss curves extended to larger model scales and compute budgets. The red pentagram indicates the projected loss for a 1.07B model with 10B training tokens, experimental results are shown as green hexagons.}
    \label{fig:Loss_combine}
\end{figure*}

\subsection{Models}
In our experiments, we employ the Cross-DiT architecture~\cite{chen2023pixart}, an efficient diffusion transformer model that incorporates cross-attention module to integrate text conditions. This architecture is optimized for high-quality image/video generation at reduced computational costs. Our model setup includes:
\begin{itemize}
    \item VAE\cite{opensora} for encoding, PixArt-XL-2-512x512~\cite{chen2023pixart} for initializing, and T5 for text encoding.
    % \item 8x downsampling with 2x2 patchify.
    \item Input sequences of 17 frames with a resolution of 256x256.
\end{itemize}

\subsection{Datasets}
We utilize the Panda-70M dataset~\cite{chen2024panda}. A test subset of 2000 samples is randomly selected for validation.

\subsection{Main Results}
We summarize the key results of our video diffusion transformers. Firstly, the fitting results for the optimal hyperparameters (i.e., learning rate and batch size) across different model sizes and training tokens are:

\vspace{-2mm}
\begin{equation}
    B_{\text{opt}} = \alpha_B T^{\beta_B} N^{\gamma_B}
    \label{eq:Bopt}
\end{equation}
\vspace{-5mm}

\begin{equation}
    \eta_{opt} = \alpha_\eta T^{\beta_\eta} N^{\gamma_\eta}
    \label{eq:lr_opt}
\end{equation}
% \vspace{-5mm}
\begin{table}[h!]
    \centering
    \begin{tabular}{|c|c|c|c|}
        \hline
        \textbf{Parameter for \( B_{opt} \)} & \( \alpha_B \) & \( \beta_B \) & \( \gamma_B \) \\
        \hline
        % 17.0287
        \textbf{Value} & $2.1797 \times 10^4$ & 0.8080 & 0.1906 \\
        \hline
        \textbf{Parameter for \( \eta_{opt} \)} & \( \alpha_\eta \) & \( \beta_\eta \) & \( \gamma_\eta \) \\
        \hline
        %1.06898590e-04, -2.39119923e-01, -3.64856688e-01
        % 1.16587480e-04, -1.09279072e-01, -2.65063080e-01
        \textbf{Value} & 0.0002 & -0.0453 & -0.1619 \\
        \hline
    \end{tabular}
    \caption{Fitting Results for \( B_{\text{opt}} \) and \( \eta_{\text{opt}} \)}
    \label{tab:fitting_results}
\end{table}

The constant term $\alpha_B = 2.1797 \times 10^4$ predicts tokens per batch. In our experiments, $\alpha_B = 17.0287$ for samples, with exponents unchanged.

Based on the optimal learning rate, we fit the validation loss for any model size and training tokens (~\Cref{tab:fitting_loss_1}).

\vspace{-2mm}
\begin{equation}
    L(T, N) = \left(\frac{T_c}{T}\right)^{\alpha_T} + \left(\frac{N_c}{N}\right)^{\alpha_N} + L_\infty
    \label{eq:loss_1}
\end{equation}
\vspace{-2mm}

\begin{table}[h!]
    \centering
    \renewcommand{\arraystretch}{1.5}  % Adjust row height
    \setlength{\tabcolsep}{4pt}  % Adjust column spacing
    \begin{tabular}{|c|c|c|c|c|c|}
        \hline
        \textbf{Parameter} & \( T_{\text{c}} \) & \( \alpha_T \) & \( N_{\text{c}} \) & \( \alpha_N \) & \( L_{\infty} \) \\
        \hline
        % 0.02354477, 0.41826788, 0.00389306, 0.29349145, 0.61830088
        \textbf{Value} & 0.0373 & 0.2917 & 0.0082 & 0.3188 & 0.4856 \\
        \hline
    \end{tabular}
    \caption{Fitting Results for \(L(T, N)\)}
    \label{tab:fitting_loss_1}
\end{table}

The optimal model size and training tokens for a fixed compute budget are given by:
\vspace{-2mm}
\begin{equation}
    N_{\text{opt}} = 0.8705 \cdot C^{0.4294}
    \label{eq:Nopt_predict_1}
\end{equation}
\vspace{-5mm}

\begin{equation}
    T_{\text{opt}} = \frac{4}{3\left(7 + \frac{n_{ctx}}{d}\right)} \cdot C^{0.5706}
    \label{eq:Topt_predict}
\end{equation}
\vspace{-2mm}

\section{Proof of Key Results}
\label{appendix:proof}
\subsection{Upper Bound of Stochastic Gradient}
\label{appendix:upper_bound}
For convenience, we restate the two assumptions here:
\begin{equation}
    \mathbb{E}[ g_k \mid \mathcal{G}_k^B] = G(\theta_k) 
    \label{eq:unbiased}
\end{equation}
\begin{equation}
    \mathbb{E}[\|g_k - G(\theta_k) \|^2 \mid \mathcal{G}_k^B] \leq \sigma_B^2 = \frac{\sigma^2}{B}
    \label{eq:variance}
\end{equation}
The assumptions indicates that, the stochastic gradient \( g_k\) is an unbiased estimate of \( G(\theta_k) \), and the variance is bounded by \( \sigma_B^2 \). Using these two assumptions we get:
\begin{align}
    & \mathbb{E}[\|g_k\|^2 \mid \mathcal{G}_k^B] \notag \\
    &= \mathbb{E}[\|g_k - G(\theta_k) + G(\theta_k)\|^2 \mid \mathcal{G}_k^B] \notag \\
    &= \|G(\theta_k)\|^2 + \mathbb{E}[\|g_k - G(\theta_k)\|^2 \mid \mathcal{G}_k^B] \notag \\
    &\leq \|G(\theta_k)\|^2 + \sigma_B^2
    \label{eq:convergence_analysis}
\end{align}
where the second equality holds due to \Cref{eq:unbiased} and the last inequality holds due to \Cref{eq:variance}.

\subsection{Convergence Rate of Mini-Batch SGD}
Since \( L(\theta_k) \) is \( L \)-smooth ,we have
\begin{align}
    & \mathbb{E}[L(\theta_{k+1}) \mid \mathcal{G}_k^B] \notag\\
    &\leq L(\theta_k) + \mathbb{E}[\langle G(\theta_k), \theta_{k+1} - \theta_k \rangle \mid \mathcal{G}_k^B] \notag\\
    &\quad + \frac{L}{2} \mathbb{E}[\|\theta_{k+1} - \theta_k\|^2 \mid \mathcal{G}_k^B] \notag\\
    &= L(\theta_k) - \eta \mathbb{E}[\langle G(\theta_k), g_k \rangle \mid \mathcal{G}_k^B] \notag\\
    &\quad + \frac{L \eta^2}{2} \mathbb{E}[\|g_k\|^2 \mid \mathcal{G}_k^B] \notag\\
    &\leq L(\theta_k) - \eta \left(1 - \frac{L \eta}{2}\right) \|G(\theta_k)\|^2 + \frac{L \eta^2 \sigma_B^2}{2} \notag\\
    &\leq L(\theta_k) - \frac{\eta}{2} \|G(\theta_k)\|^2 + \frac{L \eta^2 \sigma_B^2}{2}
    \label{eq:Converge}
\end{align}
By taking expectations over the filtration \( \mathcal{G}_k^B \), we have
\begin{align}
    \mathbb{E}[L(\theta_{k+1})] &\leq \mathbb{E}[L(\theta_k)] - \frac{\eta}{2} \mathbb{E}[\|G(\theta_k)\|^2] + \frac{L \eta^2 \sigma_B^2}{2}
\end{align}
This is equivalent to
\begin{align}
    \mathbb{E}[\|G(\theta_k)\|^2] &\leq \frac{2}{\eta} \left( \mathbb{E}[L(\theta_k)] - \mathbb{E}[L(\theta_{k+1})] \right) + L \eta \sigma_B^2
\end{align}
Taking the average over \( k = 0, 1, \dots, K \), we have
\begin{align}
    \frac{1}{K+1} \sum_{k=0}^K \mathbb{E}[\|G(\theta_k)\|^2] &\leq \frac{2 (L(\theta_0) - L^\star)}{\eta (K+1)} + L \eta \sigma_B^2
\end{align}
as required in the main text.
\subsection{Stepwise Loss of Mini-Batch SGD}
We approximate \( G(\theta_k) \) using a batch of \( B \) samples:
\begin{equation}
  g_k = \frac{1}{B} \sum_{b=1}^{B} G_{\text{est}}(\theta_k,\xi_k^{(b)}) \quad,\xi_k^{(b)}\sim \rho
  \label{eq:Gest}
\end{equation}
Following~\cite{mccandlish2018empirical}, the estimated gradient is unbiased, and its variance decreases inversely with the batch size \( B \) (see~\Cref{eq:unbiased1,eq:Gest_var}). 
% (see~\Cref{eq:unbiased1,eq:Gest_var}). 

With the Hessian matrix \( H_k \) representing the second derivatives of \( L(\theta_k) \) with respect to \( \theta_k \), the change in loss is approximately:
\begin{equation}
    L(\theta_k - \eta g_k) \approx L(\theta_k) - \eta G(\theta_k)^\top g_k + \frac{1}{2} \eta^2 g_k^\top H_k g_k
    \label{eq:loss_lg}
\end{equation}
To obtain a more stable estimate, we compute the expectation over multiple updates:
\begin{align}
    \mathbb{E} \left[ L(\theta_k - \eta g_k) \right] & \approx L(\theta_k) - \eta \|G(\theta_k)\|^2 \\ \nonumber
    &+ \frac{1}{2} \eta^2 \left( G(\theta_k)^\top H_k G(\theta_k) + \frac{\text{tr}(H_k \Sigma_{k})}{B} \right)
    \label{eq:lossupdate}
\end{align}
This allows us to express the expected loss change per update step as follows:
\begin{align}
    \Delta L_k &= \mathbb{E} \left[ L(\theta_k - \eta g_k) \right] - L(\theta_k) \\ \nonumber
    & \approx - \eta \|G(\theta_k)\|^2 + \frac{1}{2} \eta^2 \left( G(\theta_k)^\top H_k G(\theta_k) + \frac{\text{tr}(H_k \Sigma_k) }{B} \right)
    \label{eq:lossstep}
\end{align}

%-------------------------------------------------------------------------
\begin{figure*}[h]
    \centering
    % 第一张子图
    \begin{subfigure}[b]{0.31\textwidth}
        \centering
        \includegraphics[width=\textwidth]{fig/fixed_Nopt_exp.pdf}
        \caption{Empirical Loss vs. \( N \)}
        \label{fig:FixedHp_Nopt_exp}
    \end{subfigure}
    \hfill
    % 第二张子图
    \begin{subfigure}[b]{0.31\textwidth}
        \centering
        \includegraphics[width=\textwidth]{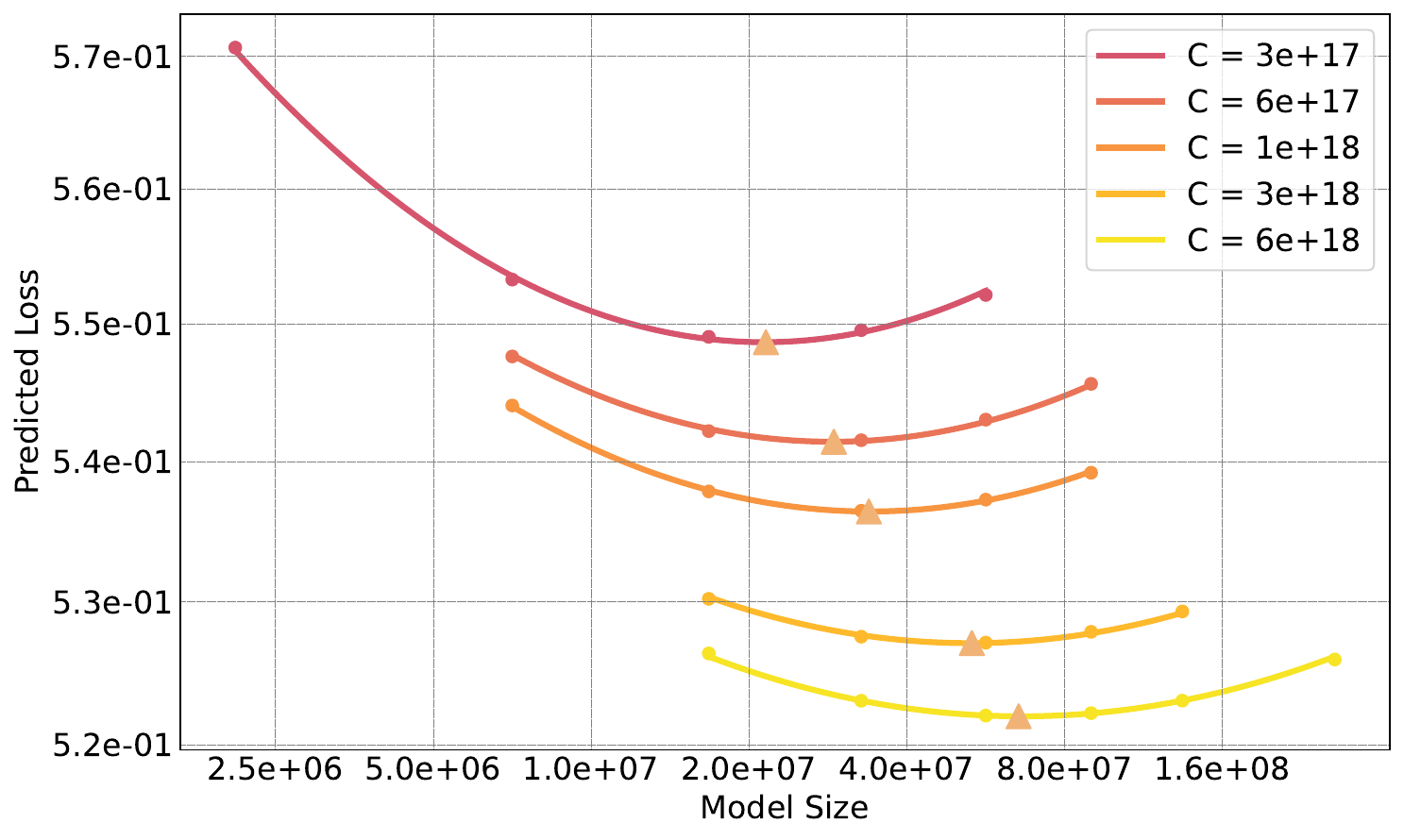}
        \caption{ Predicted Loss vs. \( N \)}
        \label{fig:FixedHp_Nopt_fit}
    \end{subfigure}
    \hfill
    % 第三张子图
    \begin{subfigure}[b]{0.3\textwidth}
        \centering
        \includegraphics[width=\textwidth]{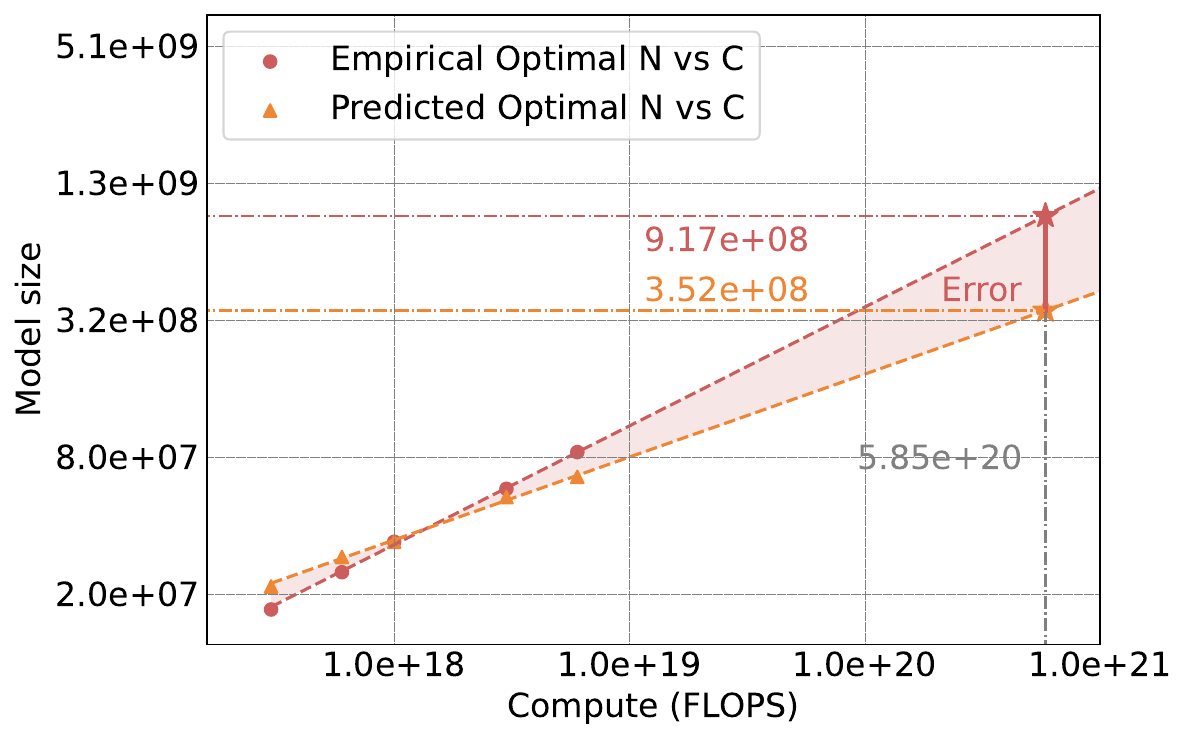}
        \caption{Optimal model scaling}
        \label{fig:FixedHp_Nopt}
    \end{subfigure}

    \caption{Empirical and predicted optimal model size on fixed suboptimal batch size and learning rate. \textbf{Left}: Empirical loss as a function of model size $N$ for various compute budgets $C$, with a parabolic fit to identify minimum loss points. \textbf{Middle}: Predicted loss across model sizes, using \Cref{eq:loss-fixed} to predict loss for different values of $N$. \textbf{Right}: Optimal model scaling with compute budgets, comparing empirical results (circles) and predicted results (triangles), confirming the accuracy of \Cref{eq:loss-fixed} for predicting the optimal model size.}
    \label{fig:fixedNopt_combine}
\end{figure*}

\begin{table*}[h]
\setlength\tabcolsep{1pt} 
\renewcommand{\arraystretch}{1.2} % Increases line spacing for readability
\centering
\begin{tabular}{|c|c|c|}
\hline
\textbf{Operation} & \textbf{Parameters} & \textbf{FLOPs} \\ \hline
Self-Attention:QKV & $3n_{\text{layer}}d^2$ & $2n_{\text{layer}}3d^2$ \\ \hline
Self-Attention:No Mask & -- & $4n_{\text{layer}}n_{ctx}d$ \\ \hline
Self-Attention:Project & $ n_{\text{layer}}d^2 $ & $2n_{\text{layer}}d^2$ \\ \hline
Cross-Attention:Q & $n_{\text{layer}}d^2$ & $2n_{\text{layer}}d^2$ \\ \hline
Cross-Attention:KV & $2n_{\text{layer}}d^2$ & $2n_{\text{layer}}2(n_{text}/n_{ctx})d^2$ \\ \hline
Cross-Attention:No Mask & -- & $4n_{\text{layer}}n_{text}d$ \\ \hline
Cross-Attention:Project & $n_{\text{layer}}d^2$ & $2n_{\text{layer}}d^2$ \\ \hline
FeedForward(SwiGLU) & $n_{\text{layer}}3dd_{ff}=n_{\text{layer}}8d^2$ & $16n_{\text{layer}}d^2$ \\ \hline
Total & $16n_{\text{layer}}d^2$ & $C=3C_{forward} = 3*(\frac{7+n_{ctx}/d}{4} N)$ \\ \hline
\end{tabular}
\caption{Parameter counts and compute estimates for the Cross-DiT model. The input dimensions are $f \times h \times w$, where $f$ is the number of frames, and $h$ and $w$ are the height and width of each frame, respectively. To ensure consistency across models of varying sizes, we maintain proportional scaling in both model width ($d$) and depth ($n_{\text{layer}}$), with $d / n_{\text{layer}} = 128$ and the number of attention heads equal to the number of layers.}
\label{table:crossdit-compute}
\end{table*}

\section{Experimental Conclusions with Fixed Suboptimal Hyperparameters}
\label{appendix:experiment}
To demonstrate that using optimal hyperparameters can yield more accurate and robust performance predictions, we simply fixed the parameters at \( B = 128 \), \( \eta = 2.5313 \times 10^{-4} \).
\begin{equation}
    L(T, N) = \left(\frac{T_c}{T}\right)^{\alpha_T} + \left(\frac{N_c}{N}\right)^{\alpha_N} + L_\infty
    \label{eq:loss-fixed}
\end{equation}

\begin{table}[t!]
    \setlength{\tabcolsep}{4pt}  % 调整列之间的间距
    \centering
    \begin{tabular}{|c|c|c|c|c|c|}
        \hline
        \textbf{Parameter} & \( T_{\text{c}} \) & \( \alpha_T \) & \( N_{\text{c}} \) & \( \alpha_N \) & \( L_{\infty} \) \\
        \hline
        \textbf{Value} & 0.0541 & 0.2515 & 0.0052 & 0.4101 & 0.4783\\
        \hline
    \end{tabular}
    \label{tab:fitting_loss_ap1}
    \caption{Fitting results for \( L(T,N) \) on fixed suboptimal hyperparameters}
\end{table}
We used the same hyperparameters as in the fitting experiment to test the 1.07B model on 10B training tokens (\Cref{fig:fixed-Loss}).

To evaluate the model's fitting performance on the data points, we calculate the mean squared error (MSE) \Cref{eq:MSE} between the fitted values and the actual data points.
\begin{equation}
    \text{MSE} = \frac{1}{n} \sum_{i=1}^{n} (y_i - \hat{y}_i)^2
    \label{eq:MSE}
\end{equation}

With fixed hyperparameters, the MSE of the fitted results is \(4.31 \times 10^{-7}\), while using optimal hyperparameters reduces it to \(2.35 \times 10^{-7}\), a $45.5\%$ improvement. This means the optimal hyperparameters are more effective for accurately capturing the model's power-law performance. 

To explore the relationship between optimal model size and compute budget, we fixed the compute budget to get the IsoFLOPs curve from~\Cref{eq:loss-fixed} and validated it through experiments (~\Cref{fig:fixedNopt_combine}).

\begin{equation}
    \hat{N}_{\text{opt}} =  9.5521 \cdot C^{0.3643}
    \label{eq:Nopt_predict}
\end{equation}

\begin{equation}
    N_{\text{opt}} = 0.0130 \cdot C^{0.5224}
    \label{eq:Nopt_exp}
\end{equation}

\Cref{eq:Nopt_predict} differs from \Cref{eq:Nopt_exp} by an absolute error of $0.1581$, or $30.26\%$, indicating a significant discrepancy, larger than the one observed in the fitting results under optimal hyperparameters. This discrepancy stems from the poor fit of \Cref{eq:loss-fixed}, particularly for lower compute budgets.

\section{Parameters and Compute}
\label{appendix:Compute}
Based on the Diffusion Transformer (DiT)~\cite{peebles2023dit},
Cross-DiT architecture incorporate cross-attention modules to inject text conditions and streamline the computation-intensive class-condition branch to improve efficiency~\cite{chen2023pixartalpha}.

To analyze the computational complexity of the CrossDiT model, we consider the parameter counts and the number of floating-point operations (FLOPs) required for a forward pass. ~\Cref{table:crossdit-compute} summarizes the parameters and compute budget for each operation within the architecture. These operations include self-attention and cross-attention, as well as feed-forward layers implemented with SwiGLU activation. The total compute cost \( C \) is derived by summing the contributions from these components. Notably, the total parameter count scales with the number of layers (\( n_\text{layer} \)) and the model width (\( d \)), ensuring proportional scaling across models of varying sizes. For consistency, \( d / n_\text{layer} = 128 \), and the number of attention heads is equal to \( n_\text{layer} \).

\section{Precise Scaling Laws in Image Generation}
\label{appendix:t2i}
The image generation can be seen as a degraded version of video generation, where the video frames are reduced to a single frame. To demonstrate that our scaling law approach also applies to diffusion transformer-based image generation, we conducted experiments using the same setup as in video generation, but fixed the number of generated frames to 1 for image generation.

Following the method in main text, we fit $\eta_{opt}(N,T)$, $B_{opt}(N,T)$. The fitting results are:

\begin{table}[h!]
    \centering
    \begin{tabular}{|c|c|c|c|}
        \hline
        \textbf{Parameter for \( B_{opt} \)} & \( \alpha_B \) & \( \beta_B \) & \( \gamma_B \) \\
        \hline
        % 5.66241187e+04, 1.49531689e-01, 3.77544648e-02
        \textbf{Value} & $5.6624 \times 10^4$ & 0.1495 & 0.0378 \\
        \hline
        \textbf{Parameter for \( \eta_{opt} \)} & \( \alpha_\eta \) & \( \beta_\eta \) & \( \gamma_\eta \) \\
        \hline
        % 1.43239029e-04, -1.86757500e-01, -2.39554012e-01
        \textbf{Value} & 0.0001 & -0.1868 & -0.2396 \\
        \hline
    \end{tabular}
    \caption{Fitting results for \( B_{\text{opt}} \) and \( \eta_{\text{opt}} \) for image generation}
    \label{tab:fitting_results}
\end{table}

Then, based on the optimal hyperparameters, we get the result of $L(N,T)$.
\begin{equation}
    L(T, N) = \left(\frac{T_c}{T}\right)^{\alpha_T} + \left(\frac{N_c}{N}\right)^{\alpha_N} + L_\infty
    \label{eq:loss-fixed}
\end{equation}

\begin{table}[h!]
    \setlength{\tabcolsep}{4pt}  % 调整列之间的间距
    \centering
    \begin{tabular}{|c|c|c|c|c|c|}
        \hline
        \textbf{Parameter} & \( T_{\text{c}} \) & \( \alpha_T \) & \( N_{\text{c}} \) & \( \alpha_N \) & \( L_{\infty} \) \\
        \hline
        \textbf{Value} & 0.0235 & 0.4183 & 0.0039 & 0.2935 & 0.6183\\
        \hline
    \end{tabular}
    \label{tab:fitting_loss_t2i}
    \caption{Fitting results for \( L(T,N) \) for image generation}
\end{table}
We extrapolated the model size to 1.07B and trained with 2B training tokens. The predicted loss was $0.6414$, the actual loss was $0.6340$, resulting in an error of $1.167\%$, which can also make accurate predictions for image generation.

\end{document}